\documentclass[11pt, letterpaper]{article}
\pdfoutput=1
\usepackage[authoryear]{natbib}
\usepackage{amsmath,amssymb,amsfonts,amsthm,bbm}
\usepackage{epic,eepic,epsfig,longtable}
\usepackage{multirow,verbatim}
\usepackage{array}

\usepackage{epsfig}
\usepackage{setspace}
\usepackage{pb-diagram}
\usepackage{fancyhdr}
\usepackage{graphicx}
\usepackage[title]{appendix}
\usepackage{listings}
\usepackage{longtable}
\usepackage{url}
\usepackage{lineno}
\usepackage{mathrsfs}
\usepackage{stmaryrd}
\usepackage{appendix}
\usepackage{algorithmic}
\usepackage{algorithm}

\newcommand\blfootnote[1]{%
  \begingroup
  \renewcommand\thefootnote{}\footnote{#1}%
  \addtocounter{footnote}{-1}%
  \endgroup
}

\makeatletter
\def\singlespace{\def\baselinestretch{1}\@normalsize}

\makeatletter
\def\singlespace{\def\baselinestretch{1}\@normalsize}

\allowdisplaybreaks
\usepackage{verbatim}
\renewcommand{\baselinestretch}{1.66}
\numberwithin{equation}{section}
\theoremstyle{plain}

\theoremstyle{definition}

\newcounter{CondCounter}

\usepackage{mystyle}
\date{\vspace{-5ex}}

\makeatother

\begin{document}

\title{Online learning in bandits with predicted context}

\author{Yongyi Guo\thanks{Department of Statistics, University of Wisconsin-Madison, Madison, WI, 53706. Email: \texttt{guo98@wisc.edu}.}
\and Ziping Xu\thanks{Department of Statistics, Harvard University, Cambridge, MA, 02138. Email: \texttt{zipingxu@fas.harvard.edu}.}
\and Susan Murphy\thanks{Department of Statistics, Harvard University, Cambridge, MA, 02138. Email: \texttt{samurphy@g.harvard.edu}.}
}


\maketitle
\onehalfspacing

\begin{abstract}
We consider the contextual bandit problem where at each time, the agent only has access to a noisy version of the context and the error variance (or an estimator of this variance). This setting is motivated by a wide range of applications where the true context for decision-making is unobserved, and only a prediction of the context by a potentially complex machine learning algorithm is available. When the context error is non-vanishing, classical bandit algorithms fail to achieve sublinear regret. We propose the first online algorithm in this setting with sublinear regret guarantees under mild conditions. The key idea is to extend the measurement error model in classical statistics to the online decision-making setting, which is nontrivial due to the policy being dependent on the noisy context observations. We further demonstrate the benefits of the proposed approach in simulation environments based on synthetic and real digital intervention datasets.
\end{abstract}

\section{Introduction}

Contextual bandits \citep{auer2002using, langford2007epoch} represent a classical sequential decision-making problem where an agent aims to maximize cumulative reward based on context information. At each round $t$, the agent observes a context and must choose one of $K$ available actions based on both the current context and previous observations. Once the agent selects an action, she observes the associated reward, which is then used to refine future decision-making. 
Contextual bandits are typical examples of reinforcement learning problems where a balance between exploring new actions and exploiting previously acquired information is necessary to achieve optimal long-term rewards. It has numerous real-world applications including personalized recommendation systems \citep{li2010contextual, bouneffouf2012contextual}, healthcare \citep{yom2017encouraging, liao2020personalized}, and online education \citep{liu2014trading, shaikh2019balancing}. 

\blfootnote{The authors have equal contribution.}

Despite the extensive existing literature on contextual bandits, in many real-world applications, the agent never observes the context \emph{exactly}. One common reason is that the true context for decision-making can only be detected or learned approximately from observable auxiliary data. For instance, consider the Sense2Stop mobile health study, in which the context is whether the individual is currently physiologically stressed \citep{battalio2021sense2stop}. A complex predictor of current stress was constructed/validated based on multiple prior studies \citep{cohen1985measuring, sarker2016finding, sarker2017markers}. This predictor was then tuned to each user in Sense2Stop prior to their quit-smoke attempt and then following the user's attempt to quit smoking, at each minute, the predictor inputs high-dimensional sensor data on the user and outputs a continuous likelihood of stress for use by the decision-making algorithm.
{In many such applications in health interventions, models using validated predictions as contexts are preferred to raw sensor data because of the high noise in these settings, and that the decision rules are interpretable so they can be critiqued by domain experts.}
The second reason why the context is not observed exactly is because of measurement error. {Contextual variables, such as user preferences in online advertising, content attributes in recommendation systems, and patient conditions in clinical trials, are prone to noisy measurement.} This introduces an additional level of uncertainty that must be accounted for when making decisions.

Motivated by the above, we consider the linear contextual bandit problem where at each round, the agent only has access to a noisy observation of the true context. Moreover, the agent has limited knowledge about the underlying distribution of this noisy observation, as in many practical applications (e.g. the above-mentioned ones). This is especially the case when this `observation' is the output of a complex machine learning algorithm. We only assume that the noisy observation is unbiased, its variance is known or can be estimated, and we put no other essential restrictions on its distribution. In healthcare applications, the estimated error variance for context variables can often be derived from data in prior studies (e.g. pilot studies). This setting is intrinsically difficult for two main reasons: First, when estimating the reward model, the agent needs to take into account the misalignment between the noisy context observation and the reward which depends on the true context. Second, even if the reward model is known, the agent may suffer from making bad decisions at each round because of the inaccurate context. 

\noindent{\textbf{Our contributions.}} We present the first online algorithm \texttt{MEB} (Measurement Error Bandit) with sublinear regret in this setting under mild conditions. \texttt{MEB} achieves $\tilde\cO(T^{2/3})$ regret compare to a standard benchmark and $\tilde\cO(T^{1/2})$ regret compare to a clipped benchmark with minimum exploration probability which is common in many applications \citep{yang2020targeting, yao2021power}. \texttt{MEB} is based on a novel approach to model estimation which removes the systematic bias caused by the noisy context observation. The estimator is inspired by the measurement error literature in statistics \citep{carroll1995measurement, fuller2009measurement}: We extend this classical method with additional tools in the online decision-making setting due to the policy being dependent on the measurement error. 

\subsection{Related work}

Our work complements several lines of literature in contextual bandits, as listed below.

\noindent\textbf{Latent contextual bandit.} In the latent contextual bandit literature \citep{zhou2016latent, sen2017contextual, hong2020latent, hong2020non, xu2021generalized, 
nelson2022linearizing, galozy2023information}, the reward is typically modeled as {jointly} depending on the latent state, the context, and the action. 
Several works \citep{zhou2016latent, hong2020latent, hong2020non, galozy2023information} assume no direct relation between the latent state and the context while setting a parametric reward model. For example, \cite{hong2020latent} assumes the latent state $s\in\cS_l$ is unknown but constant over time. 
\cite{hong2020non} assume that the latent state evolves through a Markov chain. 
\cite{xu2021generalized} sets a specific context as well as the latent feature for each action, and models the reward depending on them through a generalized linear model. Different from the aforementioned studies, we specify that the observed context is a noisy version of the latent context (which aligns with the applications we are addressing), and then leverage this structure to design the online algorithm.

In another line of work, \cite{sen2017contextual, nelson2022linearizing} consider contextual bandit with latent confounder, where the observed context influences the reward through a latent confounder variable, which ranges within a small discrete set. 
Our setting is distinct from these works in that the latent context can span an infinite (or even continuous) space.

\noindent\textbf{Bandit with noisy context.} 
In bandit with noisy context \citep{yun2017contextual, kirschner2019stochastic, yang2020multi, lin2022distributed, lin2022stochastic}, the agent has access to a noisy version of the true context and/or some knowledge of the distribution of the noisy observation. \cite{yun2017contextual, park2022worst, jose2024thompson} consider settings where 
the joint distribution of the true context and the noisy observed context is known (up to a parameter). 
Other works, such as \cite{kirschner2019stochastic, yang2020multi, lin2022stochastic}, assume that the agent knows the exact distribution of the context each time, and observes no context. By assuming a linear reward model, \cite{kirschner2019stochastic} transforms the problem into a linear contextual bandit, and obtains $\tilde\cO(d\sqrt{T})$ regret compared to the policy maximizing the expected reward over the context distribution. \cite{yang2020multi, lin2022stochastic, lin2022distributed} consider variants of the problem such as multiple linear feedbacks, multiple agents, and delayed observation of the exact context. 
Compared to these works, we consider a practical but more challenging setting where besides an unbiased noisy observation(prediction) for each context, the agent only knows the second-moment information about the distribution. This does not transform into a standard linear contextual bandit as in \cite{kirschner2019stochastic}.

\noindent\textbf{Bandit with inaccurate/corrupted context.} These works consider the setting where the context is simply inaccurate (without randomness), or is corrupted and cannot be recovered. In \cite{yang2021bandit, yang2021robust}, at each round, only an inaccurate context is available to the decision-maker, and the exact context is revealed after the action is taken. 
In \cite{bouneffouf2020online, galozy2020corrupted}, each context $\xb_t$ is completely corrupted with some probability and the corruption cannot be recovered. In \cite{ding2022robust}, the context is attacked by an adversarial agent. Because these works focus more on adversarial settings for the context observations, the application of their regret bounds to our setting generally results in a linear regret. {For example, in \cite{yang2021bandit}, the regret of Thompson sampling is $\tilde\cO(d\sqrt{T} + \sqrt{d}\sum_{t\in[T]}\|\hat\xb_t - \xb_t\|_2)$, where $\|\hat\xb_t - \xb_t\|_2$ is the error of the inaccurate context. As is typical in our setting, $\|\hat\xb_t - \xb_t\|_2$ will be non-vanishing through time, so the second term is linear in $T$.} Given the applications we consider, we can exploit the stochastic nature of the noisy context observations in our algorithm to achieve improved performance.
\subsection{Notations}
Throughout this paper, we use $[n]$ to represent the set $\{1, 2, \ldots, n\}$ for $n\in\mathbb N^+$. For $a, b\in\RR$, let $a\wedge b$ denote the minimum of $a$ and $b$. Given $d\in\mathbb N^+$, $\bI_d$ denotes the $d$-by-$d$ identity matrix, and $\mathbf{1}_d$ denotes the $d$-dimensional vector with 1 in each entry. For a vector $\bv\in\RR^d$, denote $\|\bv\|_2$ as its $\ell_2$ norm. For a matrix $\bM\in\RR^{m\times n}$, denote $\|\bM\|_2$ as its operator norm. The notation $\cO(X)$ refers to a quantity that is upper bounded by $X$ up to constant multiplicative factors, while $\tilde\cO(X)$ refers to a quantity that is upper bounded by $X$ up to poly-log factors.

\section{Measurement error adjustment to bandit with noisy context}

\subsection{Problem setting}\label{sec::setting}

We consider a linear contextual bandit with context space $\cX\subset\RR^d$ and binary action space $\cA = \{0, 1\}$\footnote{For simplicity, we state our results under the binary-action setting, which is common in healthcare \citep{trella2022reward}, economics \citep{athey2017efficient, kitagawa2018should} and other applications. However, all the results presented in this paper can be extended to the setting with multiple actions. See Appendix \ref{appendix:generalization-to-K-actions}.}. Let $T$ be the time horizon. As discussed above, we consider the setting where at each time $t\in[T]$, the agent only observes a noisy version of the context $\tilde\xb_t$ instead of the true underlying context $\xb_t$. Thus, at time $t$, the observation $o_t$ only contains $(\tilde \xb_t, a_t, r_t)$, where $a_t$ is the action and $r_t$ is the corresponding reward. We further assume that $\tilde \xb_t = \xb_t + \bepsilon_t$, where the error $\bepsilon_t$ is independent of the history $\cH_{t-1} := \{o_\tau\}_{\tau\leq t-1}$, $\EE \bepsilon_t = 0$, $\Var(\bepsilon_t) = \bSigma_{e, t}$. {Here, `e' in the subscript means `error'}. Initially we assume that $(\bSigma_{e, t})_{t\geq 1}$ is known. In Section \ref{section:estimated-Sigma-e}, we consider the setting where only estimators of $(\bSigma_{e, t})_{t\geq 1}$ are available. 
There is no restriction that the distribution of $(\bepsilon_t)_{t\geq 1}$ belongs to any known (parametric) family.
The reward $r_t = \langle\btheta_{a_t}^*, \xb_t\rangle + \eta_t$, where $\EE[\eta_t|\cH_{t-1}, \bepsilon_t, a_t] = 0$ and $(\btheta_a^*)_{a\in\cA}$ are the unknown parameters. 
It's worth noting that besides the policy, all the randomness here comes from the reward noise $\eta_t$ and the context error $\bepsilon_t$. We treat $(\xb_t)_{t\geq 1}$ as fixed throughout but unknown to the algorithm (Unlike \cite{yun2017contextual}, we don't assume $\xb_t$ are i.i.d.). Our goal at each time $t$ is to design policy $\pi_t(\cdot|\cH_{t-1}, \tilde\xb_t)\in\Delta(\cA)$ given past history $\cH_{t-1}$ and current observed noisy context $\tilde\xb_t$, so that the agent can maximize the reward by taking action $a_t\sim \pi_t(\cdot|\cH_{t-1}, \tilde\xb_t)$.

If $\bSigma_{e, t}$ is non-vanishing, standard contextual bandit algorithms are generally sub-optimal. {To see this, notice that $r_t = \langle \btheta_{a_t}^*, \xb_t\rangle + \eta_t = \langle \btheta_{a_t}^*, \tilde\xb_t\rangle + (\eta_t - \langle \btheta_{a_t}^*, \bepsilon_t\rangle)$. This means the error in the reward $r_t$ after observing the noisy context is $\eta_t - \langle \btheta_{a_t}^*, \bepsilon_t\rangle$, where $\bepsilon_t$ and $\tilde\xb_t$ are  dependent. Thus, $\EE[r_t|\tilde\xb_t, a_t]\neq \langle \btheta_{a_t}^*, \tilde\xb_t\rangle$. This is in contrast to the standard linear bandit setting, where given the true context $\xb_t$, $\EE[r_t|\xb_t, a_t]= \langle \btheta_{a_t}^*, \xb_t\rangle$, which ensures the sublinear regret of classical bandit algorithms such as UCB and Thompson sampling.} Therefore, it is necessary to design an online algorithm that adjusts for the errors $(\bepsilon_t)_{t\geq 1}$. 
We assume that the context, parameters and the reward are bounded, as below.

\begin{assumption}[Boundedness]\label{ass:boundedness}
$\forall t\in[T]$, $\|\tilde\xb_t\|_2\leq 1$; There exists a positive constant $R_\theta$ such that $\forall a\in\{0, 1\}$, $\|\btheta_a^*\|_2\leq R_\theta$; There exists a positive constant $R$ such that $\forall t\in[T]$, $|r_t|\leq R$. 
\end{assumption}


For any policy $\pi = (\pi_t)_t$, we define the (standard) cumulative regret as 
\begin{equation}\label{eq:regret}
\text{Regret}(T; \pi^*) = \sum\limits_{t\in[T]} [\EE_{a\sim \pi_t^*}\langle \btheta^*_{a}, \xb_t\rangle - \EE_{a\sim \pi_t}\langle \btheta^*_{a}, \xb_t\rangle],
\end{equation}
where 
\begin{equation}\label{eq:nonclipped-oracle-policy}
\pi_t^*(a)=
\begin{cases}
1,\quad &\text{if }a=a_t^*:=\argmax_a\langle \btheta^*_{a}, \xb_t\rangle,\\
0,\quad &\text{otherwise.}
\end{cases}
\end{equation}
We denote the standard benchmark policy $\pi^* = (\pi_t^*)_{t}$. This is summarized in the setting below.
\noindent\textbf{Setting 1.} (Standard setting) We aim to minimize $\text{Regret}(T; \pi^*)$ among the class of all policies.

In many applications including clinical trials, it's desirable to design the policy under the constraint that each action is sampled with a minimum probability $p_0$. One reason for maintaining exploration is that we can update and re-optimize the policy for future users to allow for potential non-stationarity \citep{yang2020targeting}. Additionally, keeping the exploration is also important for after-study analysis \citep{yao2021power}, especially when the goal of the analysis is not pre-specified prior to collecting data with the online algorithm. In these situations, it is desirable to consider only the policies that always maintain an exploration probability of $p_0>0$ for each arm, and compare the performance to the clipped benchmark policy $(\bar\pi_t^*)$:
\begin{equation}\label{eq:oracle-policy}
\bar \pi_t^*(a)=
\begin{cases}
1\!-\!p_0, &\text{if }a=a_t^*,\\
p_0, &\text{otherwise.}
\end{cases}
\end{equation}

This is summarized in the setting below.

\noindent\textbf{Setting 2.} (Clipped policy setting) We minimize 
\begin{equation*}
\text{Regret}(T; \bar\pi^*) = \sum\limits_{t\in[T]} [\EE_{a\sim \bar \pi_t^*}\langle \btheta^*_{a}, \xb_t\rangle - \EE_{a\sim \pi_t}\langle \btheta^*_{a}, \xb_t\rangle]
\end{equation*}
among the class of policies that explore any action with probability at least $p_0$.

In this work, we will provide policies with sublinear regret guarantees in both settings. 

\subsection{Estimation using weighted measurement error adjustment}\label{section::estimation}

In this section, we focus on learning the reward model parameters $(\btheta_a^*)_{a\in\cA}$ with data $\cH_t$ after a policy $(\pi_\tau(\cdot|\tilde\xb_\tau, \cH_{\tau-1}))_{\tau\in[t]}$ has been executed up to time $t\in[T]$. Learning a consistent model is important in many bandit algorithms for achieving low regret \citep{abbasi2011improved, agrawal2013thompson}. As we shall see in Section \ref{section:meb}, consistent estimation of $(\btheta_a^*)_{a\in\cA}$ plays an essential role in controlling the regret of our proposed algorithm.

\noindent\textbf{Inconsistency of the regularized least-squares (RLS) estimator.} UCB and Thompson sampling, the two classical bandit algorithms, both achieve sublinear regret based on the consistency of the estimator 
$
\hat\btheta_{a, RLS}^{(t)} = \big(\lambda I + \sum_{\tau\in[t]}1_{\{a_{\tau} = a\}}\xb_{\tau}\xb_{\tau}^\top\big)^{-1}\big(\sum_{\tau\in[t]}1_{\{a_{\tau} = a\}}\xb_{\tau}r_{\tau}\big)
$
under certain norms. When $\tilde\xb_\tau = \xb_\tau + \bepsilon_\tau$ is observed instead of $\xb_\tau$, the RLS estimator becomes 
$
\hat\btheta_{a, RLSCE}^{(t)} = \big(\lambda I + \sum_{\tau\in[t]}1_{\{a_{\tau} = a\}}\tilde\xb_{\tau}\tilde\xb_{\tau}^\top\big)^{-1}\big(\sum_{\tau\in[t]}1_{\{a_{\tau} = a\}}\tilde \xb_{\tau}r_{\tau}\big).
$
Here `RLSCE' means the RLS estimator with contextual error. However, when $\bSigma_{e, \tau} = \mathrm{Var}(\bepsilon_\tau)$ is non-vanishing, $\hat\btheta_{a, RLSCE}^{(t)}$ is generally no longer consistent, which may lead to bad decision-making (see Appendix \ref{appendix-failure-of-naive-estimator} for details). In the simple case where $(\xb_\tau, \bepsilon_\tau, \eta_\tau)_{\tau\in[t]}$ are i.i.d. and there is no action (i.e. set $a_\tau\equiv 0$), the inconsistency of $\hat\btheta_{a, RLSCE}^{(t)}$ is studied in the measurement error model literature in statistics \citep{fuller2009measurement, carroll1995measurement}, and is known as \emph{attenuation}.

\noindent\textbf{A naive measurement error adjustment.} A measurement error model is a type of regression model designed to accommodate inaccuracies in the measurement of regressors (i.e., instead of observing $\xb_t$, we observe $\xb_t+\bepsilon_t$ where $\bepsilon_t$ is a noise term with zero mean). As conventional regression techniques yield inconsistent estimators, measurement error models rectify this issue with adjustments to the estimator that consider these errors. In the current context when we want to estimate $\btheta_a^*$ from history $\mathcal H_t$, $(\tilde \xb_\tau)_{\tau\in[t]}$ can be viewed as regressors `measured with error', while $(r_\tau)_{\tau\in[t]}$ are dependent variables. If $(\bepsilon_\tau, \eta_\tau)_{\tau\in[t]}$ are i.i.d., $\bSigma_{e, \tau}\equiv \bSigma_e$, and there is no action (i.e. set $a_\tau\equiv0$), 
$
\hat\btheta_{0, me}^{(t)}:= \big(\frac{1}{t}\sum_{\tau\in[t]}\tilde\xb_\tau\tilde\xb_\tau^\top - \bSigma_e\big)^{-1}\big(\frac{1}{t}\sum_{\tau\in[t]}\tilde\xb_\tau r_\tau\big)
$
is a consistent estimator for $\btheta_0^*$. When multiple actions are present, a naive generalization of the above estimator, $\hat\btheta_{a, me}^{(t)}$, is 
\begin{align}
  \bigg(\sum\limits_{{\tau}\in[t]}1_{\{a_{\tau} = a\}}(\tilde\xb_{\tau}\tilde\xb_{\tau}^\top\! - \!\bSigma_e)\!\bigg)^{-1}\!\!\cdot \bigg(\!\sum\limits_{{\tau}\in[t]}1_{\{a_{\tau} = a\}}\tilde\xb_{\tau}r_{\tau}\bigg). \label{eq:naive-estimator} 
\end{align}
Unfortunately, $\hat\btheta_{a, me}^{(t)}$ is inconsistent in the multiple-action setting, even if the policy $(\pi_\tau)_{\tau\in[t]}$ is \emph{stationary and not adaptive to history}. This difference is essentially due to the interaction between the policy and the measurement error: In the classical measurement error (no action) setting, $\EE(\tilde\xb_\tau\tilde\xb_{\tau}^\top) = \xb_\tau\xb_{\tau}^\top + \bSigma_e$, and $\frac{1}{t}\sum_{\tau\in[t]}\tilde\xb_\tau\tilde\xb_{\tau}^\top - \bSigma_e$ concentrates around its expectation $\frac{1}{t}\sum_{{\tau}\in[t]}\xb_{\tau}\xb_{\tau}^\top$. Likewise, $\frac{1}{t}\sum_{{\tau}\in[t]}\tilde\xb_{\tau}r_{\tau}$ concentrates around $(\frac{1}{t}\sum_{\tau\in[t]}\xb_{\tau}\xb_{\tau}^\top)\btheta_0^*$. Combining the above parts thus yields a consistent estimator of $\btheta_0^*$. In our setting with multiple actions, however, the agent picks the action $a_\tau$ \emph{based on} $\tilde\xb_\tau$, so only certain values of $\tilde\xb_\tau$ lead to $a_\tau = a$. Therefore, for those $\tau\in[t]$ when we pick $a_\tau = a$, it's more likely that $\tilde\xb_\tau$ falls in certain regions depending on the policy, and we shouldn't expect $\EE(\tilde\xb_{\tau}\tilde\xb_{\tau}^\top)=\xb_{\tau}\xb_{\tau}^\top + \bSigma_e$ anymore. In other words, the policy creates a complicated dependence between $\tilde\xb_{\tau}\tilde\xb_{\tau}^\top$ and $1_{\{a_\tau=0\}}$ for each $\tau$, which changes the limit of $\frac1t\sum_{{\tau}\in[t]}1_{\{a_{\tau} = a\}}(\tilde\xb_{\tau}\tilde\xb_{\tau}^\top - \bSigma_e)$ (and similarly $\frac1t\sum_{{\tau}\in[t]}1_{\{a_{\tau} = a\}}\tilde\xb_{\tau}r_{\tau}$). This leads to the inconsistency of the naive estimator (See Appendix \ref{appendix-failure-of-naive-estimator} for a concrete example). In Section \ref{section:simulations}, we provide examples to show that (\ref{eq:naive-estimator}) not only deviates from the true parameters, but also leads to suboptimal decision-making.

\noindent\textbf{Our proposed estimator.}
Inspired by the above observations, for $\pi_\tau(a |\tilde\xb_\tau, \cH_{\tau-1})$ positive, we construct the following estimator for $\btheta^*_a$ given $\cH_t$, which corrects (\ref{eq:naive-estimator}) using importance weights:
\begin{align}\label{eq:proposed-estimator}
\hat\btheta_a^{(t)}:= \bigg(\hat\bSigma_{\tilde\xb, a}^{(t)} - 
\frac{1}{t}\sum\nolimits_{\tau\in[t]}\pi^{nd}_\tau(a)\bSigma_{e, \tau}\bigg)^{-1}\!\!\cdot
\hat\bSigma_{\tilde\xb, r, a}^{(t)},
\end{align}
where 
\begin{align*}
\hat\bSigma_{\tilde\xb, a}^{(t)}=\frac{1}{t}\sum_{\tau\in[t]}\frac{\pi_{\tau}^{nd}(a_\tau)}{\pi_\tau(a_\tau|\tilde\xb_\tau, \cH_{\tau-1})}1_{\{a_\tau = a\}}\tilde\xb_\tau\tilde\xb_{\tau}^\top,\\
\hat\bSigma_{\tilde\xb, r, a}^{(t)}=\frac{1}{t}\sum_{\tau\in[t]}\frac{\pi_{\tau}^{nd}(a_\tau)}{\pi_\tau(a_\tau|\tilde\xb_\tau, \cH_{\tau-1})}1_{\{a_\tau = a\}}\tilde\xb_\tau r_\tau.
\end{align*}
Here, $(\pi_\tau^{nd}(\cdot))_{\tau\in[t]}$ is a pre-specified policy (doesn't depend on $(\tilde\xb_\tau)_\tau$ or $\cH_{\tau-1}$) that can be chosen by the algorithm. We only require the following:

\begin{assumption}\label{ass:min-signal}
Let $\bw_t = \sqrt{d}\cdot \xb_t$, $\tilde \bw_t = \sqrt{d}\cdot {\tilde \xb_t}$ be scaled vectors of $\xb_t$ and $\tilde \xb_t$. Then there exist positive constants $\xi, \rho, \lambda_0$ s.t. (i) $\forall \bu\in \mathbb S^{d-1}$, $\mathbb E(\bu^\top\tilde \bw_t)^4\leq \xi$, (ii) $\EE\tilde\bw_t\tilde \bw_t^\top\preceq \nu \bI_d$, and (iii) $\forall t\geq \sqrt{T}$, $a\in\{0, 1\}$, $\frac1{t}\sum_{\tau\leq t}\pi_\tau^{nd}(a)\bw_\tau\bw_\tau^\top \succeq \lambda_{0}\bI_d$.

\end{assumption}

\begin{remark}\label{remark::ass-min-signal}
In Assumption \ref{ass:min-signal}, (i) and (ii) are standard moment assumptions. (iii) is mild. Even restricted to the choice of $\pi_\tau^{nd}(a)\equiv 1/2$, under mild conditions, the assumption can be satisfied with deterministic $(\xb_\tau)_{\tau\geq 1}$ or stochastic $(\xb_\tau)_{\tau\geq 1}$ such as an i.i.d. sequence, a weakly dependent stationary time series (e.g. multivariate ARMA process \citep{fan2017elements}), or a sequence with periodicity/seasonality with high probability (See Appendix \ref{pf:thm:theta-estimation} for details).
\end{remark}
The theorem below gives a high-probability upper bound on $\|\hat\btheta_a^{(t)}-\btheta_a^*\|_2$ (proof in Appendix \ref{pf:thm:theta-estimation}).

\begin{theorem}\label{thm:theta-estimation}
For any $t\in[T]$, denote $q_t\!:=\inf_{\tau\leq t, a\in\{0, 1\}}\pi_\tau(a|\tilde\xb_\tau, \cH_{\tau-1})$. Then under Assumptions \ref{ass:boundedness} and \ref{ass:min-signal}, there exist absolute constants $C$, $C_1$, such that as long as $q_t\geq C_1\max\{\frac{d(d+\log t)}{\lambda_0 t}, \frac{\xi(d+\log t)}{\lambda_0^2t}\}$, with probability at least $1\!-\!\frac{8}{t^2}$, $\forall a\in\{0, 1\},$ $\|\hat\btheta_a^{(t)}-\btheta_a^*\|_2$ is upper bounded by 
\begin{equation}\label{eq:thm:theta-estimation}
    \frac{C(R+\!R_\theta)d}{\lambda_0}\max\left\{\!\frac{d+\log t}{q_t t}, \frac{\sqrt{\nu} + \sqrt{\xi}}{\sqrt{d}}\sqrt{\frac{d+\log t}{q_t t}}\right\}.
\end{equation}
\end{theorem}

Unlike the existing literature on off-policy learning in contextual bandits (e.g. \cite{wang2017optimal, zhan2021off, zhang2021statistical, bibaut2021post}), the role of the importance weights here is to correct the dependence of a policy on the observed noisy context with error. The proof idea can be generalized to a large class of off-policy method-of-moment estimators, which might be of independent interest (see Appendix \ref{pf:thm:theta-estimation}).

\subsection{\texttt{MEB}: Online bandit algorithm with measurement error adjustment}\label{section:meb}

\begin{algorithm}[t]
	\caption{\texttt{MEB} (Measurement Error Bandit)}	
	\label{alg1}
	\begin{algorithmic}[1]
		\STATE \textbf{{Input}}: $(\bSigma_{e, t})_{t\in[T]}$: variance sequence of $(\bepsilon_t)_{t\in[T]}$; $(p_0^{(t)})_{t\in[T]}$: minimum selection probability at time $t\in[T]$; $T_0$: warm-up stage length
        \FOR{time $t = 1, 2, \ldots, T$}
        \IF{$t\leq T_0$}
        \STATE Set 
       $\pi_t(a|\tilde\xb_t, \!\cH_{t-1}) \!\in\![p_0^{(t)}\!, 1\!-p_0^{(t)}]$, $a\in\{0, 1\}$
        \STATE Sample $a_t\sim \pi_t(\cdot|\tilde\xb_t,\!\cH_{t-1})$
        \ELSE
        \STATE Obtain $(\hat\btheta_{a}^{(t-1)})_{a\in\{0, 1\}}$ from (\ref{eq:proposed-estimator})
        \STATE $\tilde a_t \leftarrow \argmax_{a\in\{0, 1\}}\langle\hat\btheta_{a}^{(t-1)}, \tilde\xb_t\rangle$
        \STATE Set $\pi_t(a|\tilde\xb_t,\!\cH_{t-1}):=
        \begin{cases}
        1-p_{0}^{(t)}, \thickspace\text{if }a = \tilde a_t\\
        p_{0}^{(t)}, \thickspace\text{otherwise}
        \end{cases}
        $
        \STATE Sample $a_t\sim \pi_t(\cdot|\tilde\xb_t,\!\cH_{t-1})$
        \ENDIF
        \ENDFOR
	\end{algorithmic}
\end{algorithm}

We propose \texttt{MEB} (Measurement Error Bandit), an online bandit algorithm with measurement error adjustment based on the estimator (\ref{eq:proposed-estimator}). The algorithm is presented in Algorithm \ref{alg1} and is designed for the binary-action setting, although it can be generalized to the case with multiple actions (see Appendix \ref{appendix:generalization-to-K-actions}). {For $t\leq T_0$, the algorithm is in a warm-up stage and can pick any policy such that there is a minimum sampling probability $p_0^{(t)}$ for each action (Here $p_0^{(t)}\in(0, \frac{1}{2}]$).} {For instance, the algorithm can do pure exploration with $\pi_t(a|\tilde\xb_t, \cH_{t-1})\equiv \frac12$.} For $t>T_0$, given the noisy context $\tilde\xb_t$, the algorithm computes the best action $\tilde a_t$ according to $(\hat\btheta_a^{(t-1)})_{a\in\{0, 1\}}$ calculated from (\ref{eq:proposed-estimator}). Then, it samples $\tilde a_t$ with probability $1-p_0^{(t)}$ and keeps an exploration probability of $p_0^{(t)}$ to sample the other action. In practice, we can often set $p_0^{(t)}$ to be monotonically decreasing in $t$, in which case $q_t\!=\inf_{\tau\leq t, a\in\{0, 1\}}\pi_\tau(a|\tilde\xb_\tau, \cH_{\tau-1}) = p_0^{(t)}$ for all $t\in[T]$.

Before presenting the regret analysis, we should first note that our problem is harder than a standard contextual bandit: $\xb_t$ is unknown, and only $\tilde\xb_t$ is observed. Thus, even if $(\btheta_a^*)_{a\in\{0, 1\}}$ is known, we may still perform suboptimally if $\tilde\xb_t$ is too far from $\xb_t$ so that it leads to a different optimal action. Example \ref{ex::unknown-xt-linear-regret} below shows that in general, we cannot avoid a linear regret.

\begin{example}\label{ex::unknown-xt-linear-regret}
Let $d = 1$, $(\btheta_1^*, \btheta_0^*) = (1, -1)$. $(\xb_t)_{t\in[T]}$ are drawn i.i.d. from $\{\pm0.2\}$ with equal probability. $\PP(\tilde\xb_t = 1|\xb_t=0.2) = 0.6$, $\PP(\tilde\xb_t = -1|\xb_t=0.2) = 0.4$; $\PP(\tilde\xb_t = 1|\xb_t=-0.2) = 0.4$, $\PP(\tilde\xb_t = -1|\xb_t=-0.2) = 0.6$. Intuitively, even if we know $(\btheta_a^*)_{a\in\{0, 1\}}$, there is still a constant probability at each time $t$ that we cannot make the right choice due to $\tilde \xb_t$ and $\xb_t$ having different signs, and $\xb_t$ is never known (details in Appendix \ref{pf:cor:alg1-with-theta-estimation}). This results in a $\Omega(T)$ regret.
\end{example}

Fortunately, in practice, we expect that the errors $(\bepsilon_t)_{t\in[T]}$ are relatively `small' in the sense that the optimal action (given $(\btheta_a^*)_{a\in\{0, 1\}}$) is not affected. Specifically, we assume the following:

\begin{assumption}\label{ass:small-error}
There exist a constant $\rho\in(0, 1)$ such that $\forall t\in[T]$, 
$|\langle\bm{\delta}_\theta, \bepsilon_t\rangle|\leq \rho|\langle\bm{\delta}_\theta, \xb_t\rangle|$ almost surely. Here $\bm{\delta}_\theta:= \btheta_1^* - \btheta_0^*$.
\end{assumption}

Assumption \ref{ass:small-error} ensures that the perturbation to the suboptimality gap between the two arms caused by $\bepsilon_t$ is controlled by the true suboptimality gap. In this way, given $(\btheta_a^*)_{a\in\{0, 1\}}$, the optimal action based on $\tilde\xb_t$ will not deviate too much from that based on $\xb_t$. As a special case, this assumption is satisfied with if $\forall t$, $|\langle\bm{\delta}_\theta, \xb_t\rangle|\geq {B_{e, t}}\|\bm{\delta}_\theta\|_2/{\rho}$. Here $B_{e, t}$ is an upper bound of $\|\bepsilon_t\|_2$. Assumption \ref{ass:small-error} can be further weakened to the inequalities holding with high probability (see Appendix \ref{pf:cor:alg1-with-theta-estimation}). Note that Assumption \ref{ass:small-error} only guarantees the optimal action is not affected by $(\bepsilon_t)_{t\in[T]}$ \emph{given} $(\btheta_a^*)_{a\in\{0, 1\}}$. To achieve sublinear regret, $(\btheta_a^*)_{a\in\{0, 1\}}$ still needs to be well-estimated. Thus, even with Assumption \ref{ass:small-error}, classical bandit algorithms such as UCB may still suffer from linear regret because of the inconsistent estimator $\hat\btheta_{a, RLSCE}^{(t)}$ (see Appendix \ref{appendix-failure-of-naive-estimator} for a concrete example).

We first prove the following theorem, which states that the regret of \texttt{MEB} can be directly controlled by the estimation error. In fact, this theorem holds regardless of the form or quality of the estimation procedure (i.e. in line 7 of Algorithm \ref{alg1}). The proof is in Appendix \ref{pf:cor:alg1-with-theta-estimation}.

\begin{theorem}\label{thm-regret}
Let Assumption \ref{ass:boundedness} and \ref{ass:small-error} hold.

(i) For the standard setting, Algorithm \ref{alg1} outputs a policy with $\text{Regret}(T; \pi^*)$ no more than 
$$
2T_0R_{\theta}+\frac{2}{1\!-\!\rho}\cdot\!\!\!\sum\limits_{t=T_0\!+1}^T\!\!\!\!\big(p_0^{(t)}R_\theta + \!\!\max\limits_{a\in\{0, 1\}}\!\|\hat\btheta_a^{(t-1)}-\btheta_a^*\|_2\big).
$$

(ii) For the clipped policy setting, Algorithm \ref{alg1} with the choice of $p_0^{(t)} \equiv p_0$ outputs a policy with $\text{Regret}(T; \bar\pi^*)$ no more than 
$$
2T_0R_{\theta}+\frac{2(1\!-\!2p_0)}{1\!-\!\rho}
\cdot\!\!\!\sum\limits_{t=T_0\!+1}^T\max\limits_{a\in\{0, 1\}}\|\hat\btheta_a^{(t-1)}\!-\!\btheta_a^*\|_2.
$$
\end{theorem}
The following corollary provides regret guarantees of \texttt{MEB} by combining Theorem \ref{thm:theta-estimation} and \ref{thm-regret} (proof in Appendix \ref{pf:cor:alg1-with-theta-estimation}).

\begin{corollary}\label{cor:alg1-with-theta-estimation}
Let Assumption \ref{ass:boundedness} to \ref{ass:small-error} hold. There exist universal constants $C, C'$ such that:

(i) For the standard setting, $\forall T\geq C'\max\{(1+{1}/{\lambda_0^{\frac94}})(d+\log T)^3, (\xi/\lambda_0)^{\frac94}(d+\log T)^{\frac43}\}$, with probability at least $1\!-\!\frac{16}{\sqrt{T}}$, Algorithm \ref{alg1} with the choice of $T_0 = \lceil 2dT^{\frac23}\rceil$, $p_0^{(t)} = \min\{\frac12, t^{-\frac13}\}$ outputs a policy with $\text{Regret}(T; \pi^*)$ no more than
$$
CdT^{\frac23}\left\{\frac{R_\theta}{1-\rho} + \frac{(\!\sqrt{\nu}\! +\!\! \sqrt{\xi}\! +\!\! 1)(R+R_\theta)}{(1-\rho)\lambda_0} \sqrt{1+\frac{\log T}{d}}\right\}.
$$

(ii) For the clipped policy setting, $\forall T\geq C'\max\{(d+\log T)^2/(\lambda_0p_0)^2, \xi^2/\lambda_0^4(1+\log T/d)^2\}$, with probability at least $1\!-\!\frac{16}{\sqrt{T}}$, Algorithm \ref{alg1} with the choice of $T_0 = \lceil2d\sqrt{T}\rceil$ and  $p_0^{(t)}\equiv p_0$ outputs a policy with $\text{Regret}(T; \bar\pi^*)$ no more than 
$$
CdT^{\frac12}\bigg\{\!\!R_\theta + \frac{(\!\sqrt{\nu}\! +\!\! \sqrt{\xi}\! +\!\! 1)(1\!-\!2p_0)(R\!+\!R_\theta)}{\sqrt{p_0}(1\!-\!\rho)\lambda_0}\sqrt{1\!+\!\frac{\log T}{d}}\bigg\}.
$$
\end{corollary}
Ignoring other factors, the regret upper bound is of order $\tilde\cO({d}T^{2/3})$ for the standard setting, and  $\tilde\cO(d\sqrt{T})$ for the clipped policy setting, depending on the horizon $T$ and dimension $d$.

In certain scenarios (e.g. when $d$ is large), it is desirable to save computational resources by updating the estimates of $(\btheta_a^*)_{a\in\{0, 1\}}$ less frequently in Algorithm \ref{alg1}. Fortunately, low regret guarantees can still be achieved: Suppose at each time $t$, the agent only updates the estimators according to (\ref{eq:proposed-estimator}) at selected time points $t\in\cS\subseteq[T]$ (in line 7); Otherwise, the agent simply makes decisions based on the most recently updated estimators. In Appendix \ref{pf:cor:alg1-with-theta-estimation}, we show that time points to perform the updates can be very infrequent, such as $(n^k)_{k\in\mathbb N^+}$ ($n\geq 2, n\in\mathbb N^+$), while still achieving the same rate of regret upper bound as in Corollary \ref{cor:alg1-with-theta-estimation}.

\subsection{\texttt{MEB} given estimated error variance}\label{section:estimated-Sigma-e}

In practice, the agent might not have perfect knowledge about $\bSigma_{e, t}$, the variance of the error $\bepsilon_t$. In this section, we discuss the situation where at each time $t$, the agent does not know $\bSigma_{e, t}$, and only has a (potentially adaptive) estimator $\hat\bSigma_{e, t}$ for $\bSigma_{e, t}$. This estimator may be derived from auxiliary data or outside knowledge. In this case, in Algorithm \ref{alg1}, we need to replace the estimator (\ref{eq:proposed-estimator}) with the following estimator for decision-making (i.e. in line 7 of Algorithm \ref{alg1}): 
\begin{equation}\label{eq::proposed-estimator-estimated-Sigma}
\tilde\btheta_a^{(t)}:= \big(\hat\bSigma_{\tilde\xb, a}^{(t)} - 
\frac{1}{t}\sum\nolimits_{\tau\in[t]}\pi^{nd}_\tau(a)\hat\bSigma_{e, \tau}\big)^{-1}\cdot
\hat\bSigma_{\tilde\xb, r, a}^{(t)}.
\end{equation} 
In Appendix \ref{appendix:estimated-error-variance}, we show that with this modification, the additional regret of Algorithm \ref{alg1} is controlled by 
$$d\cdot \sum\nolimits_{t=T_0+1}^T\max\nolimits_{a\in\{0, 1\}}\|\bDelta_t(a)\|_2$$ 
up to a constant depending on the assumptions. Here, for each $t$, $\bDelta_t(a):= \frac{1}{t}\sum_{\tau\in[t]}\pi_\tau^{nd}(a)(\hat\bSigma_{e, \tau} - \bSigma_{e, \tau})$ is the weighted average of the estimation errors $(\hat\bSigma_{e, \tau} - \bSigma_{e, \tau})_{\tau\in[t]}$. In practice, it is reasonable to assume that $\bDelta_t(a)$ is small so as not to significantly affect the overall regret: For example, suppose the agent gathers more auxiliary data over time so that $\|d(\hat\bSigma_{e, t} - \bSigma_{e, t})\|_2\lesssim {\sqrt{d/t}}$, then the additional regret term will be $\cO(\sqrt{dT})$ up to a constant depending on the assumptions.

\section{Simulation results}\label{section:simulations}

In this section, we complement our theoretical analyses with simulation results on a synthetic environment with artificial noise and reward models as well as a simulation environment based on the real dataset, HeartStep V1 \citep{10.1093/abm/kay067}.

\paragraph{Compared algorithms.} In both simulation environments, we compare the following algorithms: Thompson sampling (\texttt{TS}) with normal priors \citep{russo2018tutorial}, Linear Upper Confidence Bound (\texttt{UCB}) approach \citep{chu2011contextual}, \texttt{MEB} (Algorithm \ref{alg1}), and \texttt{MEB-naive} (\texttt{MEB} plugged in with the naive measurement error estimator (\ref{eq:naive-estimator}) instead of (\ref{eq:proposed-estimator})). See Appendix \ref{appendix-simulation+} for a detailed description of the algorithms.

\subsection{Synthetic environment}

We first test our algorithms on a synthetic environment. We consider a contextual bandit environment with $d = 5$, $T = 50000$. In the reward model, we set $\btheta_0^* = (5, 6, 4, 6, 4)$, $\btheta_1^* = (6, 5, 5, 5, 5)$, and $\eta_t$ drawn i.i.d. from $\mathcal{N}(0, \sigma_{\eta}^2)$. Let $(\xb_t)_{t\in[T]}$ be independently sampled from $\mathcal{N}(\bmu_{x}, \bI_d)$, where $\bmu_x = \one_d$. We further set $\bSigma_{e, t} \equiv \bSigma_e \coloneqq \bI_d / 4$ and consider independent $(\bepsilon_{t})_{t\in[T]}$ with Normal distribution
with covariance $\bSigma_{e}$. We independently generate bandit data for $n_{exp} = 100$ times, and compare among the candidate algorithms in terms of estimation quality and cumulative regret with a moderate exploration probability $p_0 = 0.2$.

\subsection{HeartStep V1 simulation environment}

We also construct a simulation environment with HeartSteps dataset. HeartSteps is a physical activity mobile health application, whose primary goal is to help the user prevent negative health outcomes and adopt and maintain healthy behaviors, for example, higher physical activity level. HeartSteps V1 is a 42-day mobile health trial \citep{dempsey2015randomised,klasnja2015microrandomized,liao2016sample}, where participants are provided a Fitbit tracker and a mobile phone application. One of the intervention components is a contextually-tailored physical activity suggestion that may be delivered at any of the five user-specified times during each day. The delivery times are roughly separated by 2.5 hours.


\paragraph{Construction of the simulated environment.} We follow the simulation setups in \cite{liao2020personalized}.  The true context at the time $t$ is denoted by $\xb_t$ with three main components $\xb_t = (I_t, Z_t, B_t)$. Here, $I_t$ is an indicator variable of whether an intervention ($A_t = 1$) is feasible (e.g. $I_t$ is 0 when the participant is driving a car, a situation where the suggestion should not be sent). $Z_t$ contains some features at time $t$. $B_t$ is the true treatment burden, which is a function of the participant's treatment history\footnote{Note this violates contextual bandits assumption and leads to an MDP. We believe this is a good setup to test the robustness of our proposed approach.}. Specifically, $B_{t+1} = \lambda B_t + 1_{\{A_{t} = 1\}}$. We assume that $(I_t)_{t\in[T]}$ and $(Z_t)_{t\in[T]}$ are sampled i.i.d with the empirical distribution from the Heartstep V1 dataset, and $(B_t)_{t\in[T]}$ is given by the aforementioned transition model.

The reward model is 
$
    r_t(\xb, a; \btheta) = \xb^{\top} \balpha + a f(\xb)^{\top} \bbeta + \eta_t,
$
where $\xb$ is the full context, $f(\xb)$ is a subset of $\xb$ that is considered to have an impact on the treatment effects, and $\btheta = (\balpha^\top, \bbeta^\top)^\top \in \mathbb{R}^9$. Here $\eta_t$ is the Gaussian noise on the reward observation, whose variance $\sigma_{\eta}^2$ is chosen to be $0.1, 1.0$, and $5.0$ respectively \citep{liao2016sample}. For a detailed list of variables in the context, see Table \ref{tab:variables} in Appendix \ref{appendix-simulation+}.

The true parameters $(\btheta_a^*)_{a\in\{0,1 \}}$ is estimated from GEE (Generalized Estimating Equations) with rewards being the log-transformed step count collected 30 minutes after the decision time.

In light of the measurement error setting in this paper, we consider an observation noise on $B_t$ for the following reasons: 1) The burden $B_t$ can be understood as a prediction of the burden level of the participant, which is particularly crucial in mobile health studies; 2) Other variables are normally believed to have low or no observation noise. Thus, we assume that the agent only observes
$
\tilde{\xb}_t = (I_t, Z_t, \tilde{B}_t),
$
where $\tilde{B}_t = B_t + \epsilon_t$ and $\epsilon_t$ is drawn i.i.d. from normal distribution with mean zero and variance $\sigma_{\epsilon}^2$.

\subsection{Results} Table \ref{tab:synthetic} (a) and (b) shows the average regret (cumulative regret divided by $T$) in both the synthetic environment and the real-data environment based on HeartStep V1. We use the same set of $\sigma_{\epsilon}^2 \in \{0.1, 1.0, 2.0\}$, while different $\sigma^2_{\eta}$ reflect the change of absolute values in coefficients in two different environments ($\sigma_{\eta}^2 = 5.0$ is the level of reward noise in HeartStep V1). \texttt{MEB} shows significantly smaller average regret compared to other baseline methods under most combinations of $\sigma^2_{\eta}$ and $\sigma_{\epsilon}^2$. 1. In certain instances, MEB-naive exhibits performance comparable to MEB. This is attributed to its ability to reduce the variance of model estimation while incurring some bias compared to MEB, rendering it a feasible alternative in practical contexts. Notably, in two extreme scenarios, UCB surpasses both MEB and MEB-naive. This is as expected, since when contextual noise is sufficiently negligible, traditional bandit algorithms are anticipated to outperform the proposed algorithms.
 An estimation error plot can be found in Appendix \ref{appendix-simulation+},
which also demonstrates that \texttt{MEB} has a lower estimation error.

\begin{table}[ht]
\caption{Average regret for both synthetic environment and real-data environment under different combinations of $\sigma_{\eta}^2$ and $\sigma_{\epsilon}^2$. The results are averages over 100 independent runs and the standard deviations are reported in the full table in Appendix \ref{appendix-simulation+}.}
\centering
\begin{subtable}[h]{0.5\textwidth}
\caption{Average regret in the synthetic environment over $50000$ steps with clipping probability $p = 0.2$. }
\begin{tabular}{ll|llll}
$\sigma_{\eta}^2$ & $\sigma_{\epsilon}^2$ & \texttt{TS} & \texttt{UCB} & \texttt{MEB} & \texttt{MEB-naive} \\
\hline
\hline
0.01 & 0.1 & 0.047 & 0.046 & \textbf{0.027} & 0.038 \\
0.1  & 0.1 & 0.047 & 0.047 & \textbf{0.026} & 0.039 \\
1.0  & 0.1 & 0.048 & 0.048 & \textbf{0.027} & 0.038 \\
0.01 & 1.0 & 0.757 & 0.647 & \textbf{0.198} & 0.371 \\
0.1  & 1.0 & 0.769 & 0.721 & \textbf{0.205} & 0.392 \\
1.0  & 1.0 & 0.714 & 0.697 & \textbf{0.218} & 0.404 \\
0.01 & 2.0 & 1.492 & 1.504 & \textbf{0.358} & 0.616 \\
0.1  & 2.0 & 1.195 & 1.333 & \textbf{0.368} & 0.584 \\
1.0  & 2.0 & 1.299 & 1.476 & \textbf{0.416} & 0.625\\
\hline
\hline
\end{tabular}
\end{subtable}

\vspace{3mm}

\begin{subtable}[h]{0.5\textwidth}
\caption{Average regret in the real-data environment over $2500$ steps with clipping probability $p = 0.2$. }
\begin{tabular}{ll|llll}
$\sigma_{\eta}^2$ & $\sigma_{\epsilon}^2$ & \texttt{TS} & \texttt{UCB} & \texttt{MEB} & \texttt{MEB-naive} \\
\hline
\hline
0.05 & 0.1 & 0.027 & 0.027 & \textbf{0.022} & 0.024 \\
0.1  & 0.1 & 0.026 & 0.024 & \textbf{0.020} & \textbf{0.020} \\
5.0  & 0.1 & 1.030 & \textbf{0.743} & 0.831 & 1.173 \\
0.05 & 1.0 & 0.412 & 0.408 & 0.117 & \textbf{0.112} \\
0.1  & 1.0 & 0.309 & 0.316 & \textbf{0.085} & 0.087 \\
5.0  & 1.0 & 1.321 & \textbf{0.918} & 1.458 & 1.322 \\
0.05 & 2.0 & 0.660 & 0.634 & \textbf{0.144} & 0.148 \\
0.1  & 2.0 & 0.740 & 0.704 & \textbf{0.151} & 0.155 \\
5.0  & 2.0 & 1.585 & 2.415 & 1.577 & \textbf{1.436}
\end{tabular}
\end{subtable}
\label{tab:synthetic}
\end{table}

\section{Discussion and conclusions}

We propose a new algorithm, \texttt{MEB}, which is the first algorithm with sublinear regret guarantees in contextual bandits with noisy context, where we have limited knowledge of the noise distribution. This setting is common in practice, especially where only predictions for unobserved context are available.
\texttt{MEB} leverages the novel estimator (\ref{eq:proposed-estimator}), which extends the conventional measurement error adjustment techniques by considering the interplay between the policy and the measurement error. 

\noindent\textbf{Limitations and future directions.} Several questions remain for future investigation. First, is $\tilde\cO(T^{2/3})$ the optimal rate of regret compared to the standard benchmark policy (\ref{eq:oracle-policy}), as in some other bandits with semi-parametric reward model (e.g. \cite{xu2022towards})? Providing lower bounds on the regret helps us understand the limit of improvement in the online algorithm. Second, we assume that the agent has an unbiased prediction of the true context. It is important to understand how biased predictions affect the results. 
Last but not least, it's interesting to see our method can be extended to more complicated decision-making settings (e.g. Markov decision processes).

\newpage 
\bibliographystyle{ims}
\bibliography{rl_aistats}
\newpage

\onecolumn


\appendix

\section{Additional details for simulation studies}\label{appendix-simulation+}

\subsection{Compared algorithms}
In both simulation environments, we compare the following algorithms: Thompson sampling (\texttt{TS}, see details in Algorithm \ref{alg:TS}) given normal priors \citep{russo2018tutorial}, Linear Upper Confidence Bound (\texttt{UCB}, see details in Algorithm \ref{alg:UCB}) approach \citep{chu2011contextual}, \texttt{MEB} (Algorithm \ref{alg1}), and \texttt{MEB-naive} (\texttt{MEB} plugged in with the naive measurement error estimator (\ref{eq:naive-estimator}) instead of (\ref{eq:proposed-estimator})). To make a fair comparison between algorithms, we use the same regularization parameter $l = 1$ for all algorithms. The hyper-parameter $\rho$, $C$ is set to be $\sigma_{\eta}^2$ for all results for \texttt{TS} and \texttt{UCB}.

\begin{algorithm}[H]
	\caption{Linear Thompson Sampling (\texttt{TS})}	
	\label{alg:TS}
	\begin{algorithmic}[1]		
\STATE \textbf{{Input}}: $T$: total number of steps; $\rho$: variance of $(\eta_t)_{t\in[T]}$; $l$: prior variance; $p_0$: minimum selection probability; $\bmu_{0, a} = \mathbf{0}$ and $\bSigma_{0, a} = l \bI$
        \FOR{time $t = 1, 2, \ldots, T$}
        \STATE Observe $\tilde \bx_t = \bx_t + \bepsilon_t$
        \STATE Generate posterior sample $(\tilde\btheta_{t, a})_{a\in\cA}$ from $\mathcal{N}(\bmu_{t-1, a}, \bSigma_{t-1, a})$
        \STATE Set $\tilde a_t \leftarrow \argmax_{a\in\cA}\langle\tilde\btheta_{t, a}, \tilde\xb_t\rangle$
        \STATE Sample $a_t\sim \pi_t(\cdot|\tilde\xb_t, \cH_{t-1})$, where $\pi_t(a|\tilde\xb_t, \cH_{t-1}):=
        \begin{cases}
        1-(K-1)p_{0}, \quad\text{if }a = \tilde a_t\\
        p_{0}, \quad\text{otherwise}
        \end{cases}
        $
        \STATE Observe reward $r_t = \langle \btheta_{a_t}^\star, \bx_t \rangle + \eta_t$
        \STATE Set $\bV_{t, a} = l\bI + \sum_{t'= 1}^t \mathbf{1}_{\{a_{t'} = a\}} \tilde \bx_t\tilde \bx_t^{\top}$, $\bb_t = \sum_{t' = 1}^t \mathbf{1}_{\{a_{t'} = a\}} \tilde \bx_t r_t$
        \STATE Update $\bmu_{t, a} = \bV_{t, a}^{-1} \bb_{t, a}$ and $\bSigma_{t, a} = \rho \bV_{t, a}^{-1}$
        \ENDFOR
	\end{algorithmic}
\end{algorithm}

\begin{algorithm}[H]
	\caption{Linear UCB (\texttt{UCB})}	
	\label{alg:UCB}
	\begin{algorithmic}[1]		
\STATE \textbf{{Input}}: $T$: total number of steps; $C > 0$; $l$: regularization; $p_0$: minimum selection probability; $\bV_{0, a} = l \bI$ and $\bb_{0, a} = \mathbf{0}$
        \FOR{time $t = 1, 2, \ldots, T$}
        \STATE Observe $\tilde \bx_t = \bx_t + \bepsilon_t$
        \STATE Estimate $\hat\btheta_{t, a} = \bV_{t-1, a}^{-1} \bb_{t-1, a}$
        \STATE Set $\hat\mu_a = \langle \hat \btheta_{t, a}, \tilde \bx_{t} \rangle + C\sqrt{\tilde\bx_{t}^{\top} \bV_{t, a}^{-1} \tilde \bx_t}$
        \STATE Set $\tilde a_t \leftarrow \argmax_{a\in\cA} \hat \mu_a$
        \STATE Sample $a_t\sim \pi_t(\cdot|\tilde\xb_t, \cH_{t-1})$, where $\pi_t(a|\tilde\xb_t, \cH_{t-1}):=
        \begin{cases}
        1-(K-1)p_{0}, \quad\text{if }a = \tilde a_t\\
        p_{0}, \quad\text{otherwise}
        \end{cases}
        $
        \STATE Observe reward $r_t = \langle \btheta_{a_t}^\star, \bx_t \rangle + \eta_t$
        \STATE Set $\bV_{t, a} = l\bI + \sum_{t'= 1}^t \mathbf{1}_{\{a_{t'} = a\}} \tilde \bx_t\tilde \bx_t^{\top}$, $\bb_t = \sum_{t' = 1}^t \mathbf{1}_{\{a_{t'} = a\}} \tilde \bx_t r_t$
        \ENDFOR
	\end{algorithmic}
\end{algorithm}

We further compare with robust linear UCB (Algorithm 1 in \cite{ding2022robust}) that is shown to achieve minimax rate for adversarial linear bandit.

\subsection{Additional details for HeartStep V1 study}
Table \ref{tab:variables} presents the list of variables to include in the reward model and in the feature construction for algorithms. Recall that our reward model is $r_t(\bx, a, \btheta) = \bx^{\top} \balpha + a f(\bx)^{\top} \bbeta + \eta_t$. All the variables are included in $\bx$ while only those considered to have an impact on treatment effect will be included in $f(\bx)$.
\begin{table}[ht]
    \centering
    \caption{List of variables in HeartSteps V1 study.}
    \begin{tabular}{l|lc}
        Variable & Type & Treatment? \\
        \hline
        Availability ($I_t$) & Discrete & No \\
        Prior 30-minute step count & Continuous & No \\
        Yesterday's step count & Continuous & No \\
        Prior 30-minute step count & Continuous & No \\
        Location & Discrete & Yes \\
        Current temperature & Continuous & No \\
        Step variation level & Discrete & Yes \\
        Burden variable ($B_t$) & Continuous & Yes
    \end{tabular}
    \label{tab:variables}
\end{table}

\subsection{Additional results on estimation error}

\begin{figure}[H]
     \begin{flushleft}
    {\quad \quad  \tiny\quad $\sigma_{\eta}^2 = 0.01 $} \hspace{35mm} {\tiny\quad $\sigma_{\eta}^2 = 0.1 $} \hspace{35mm} {\tiny\quad $\sigma_{\eta}^2 = 1.0 $}
    \end{flushleft}
    \centering
    \begin{subfigure}{1\textwidth}
    \rotatebox{90}{\tiny\quad $\sigma_{\epsilon}^2 = 0.1$}
    \includegraphics[width = 0.3\textwidth]{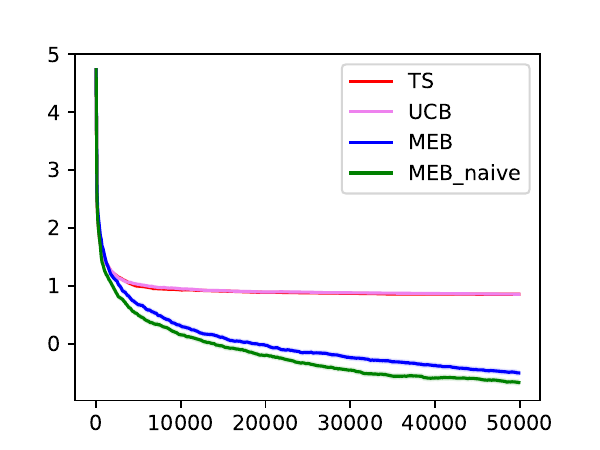}   
    \includegraphics[width = 0.3\textwidth]{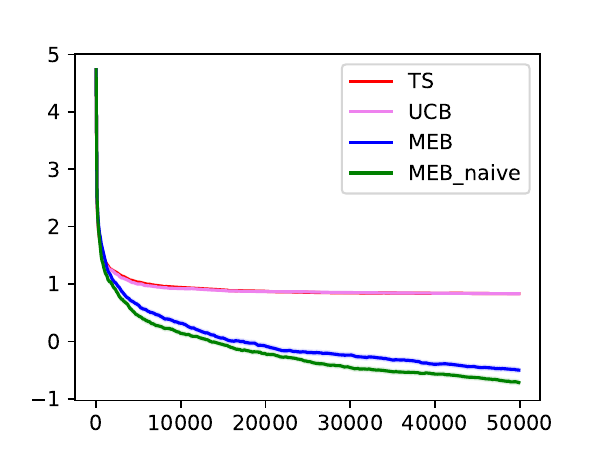}   
    \includegraphics[width = 0.3\textwidth]{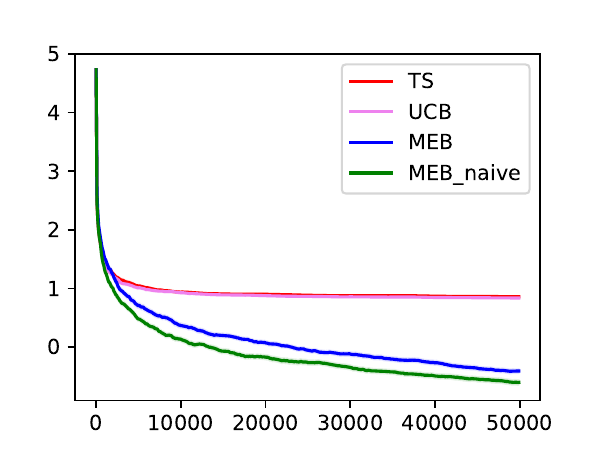}   
    \end{subfigure}
    \begin{subfigure}{1\textwidth}
    \rotatebox{90}{\tiny\quad $\sigma_{\epsilon}^2 = 1.0$}
    \includegraphics[width = 0.3\textwidth]{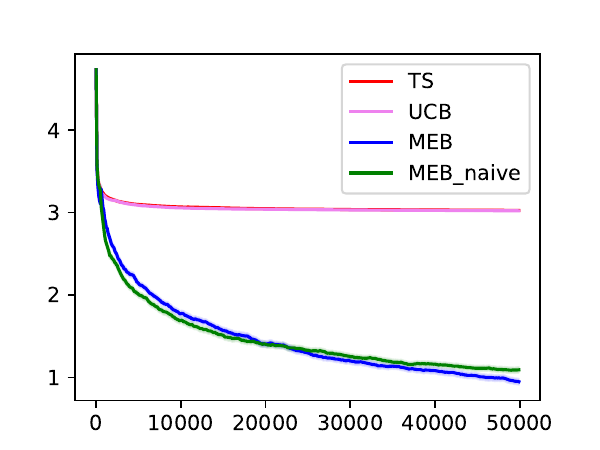}   
    \includegraphics[width = 0.3\textwidth]{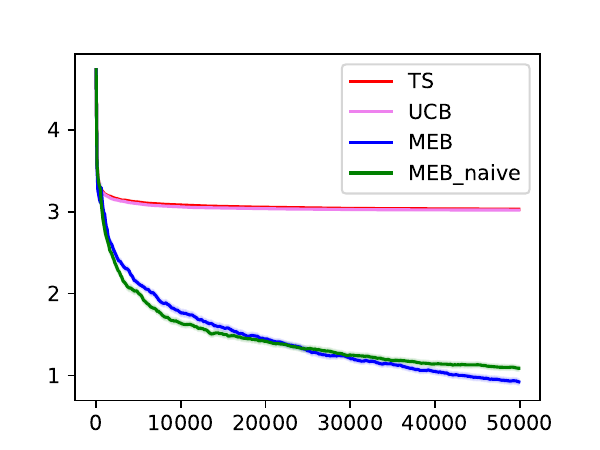}   
    \includegraphics[width = 0.3\textwidth]{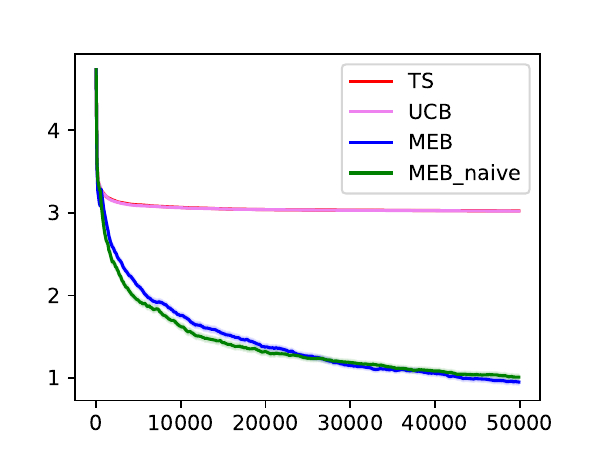}   
    \end{subfigure}
    \begin{subfigure}{1\textwidth}
    \rotatebox{90}{\tiny\quad $\sigma_{\epsilon}^2 = 2.0$}
    \includegraphics[width = 0.3\textwidth]{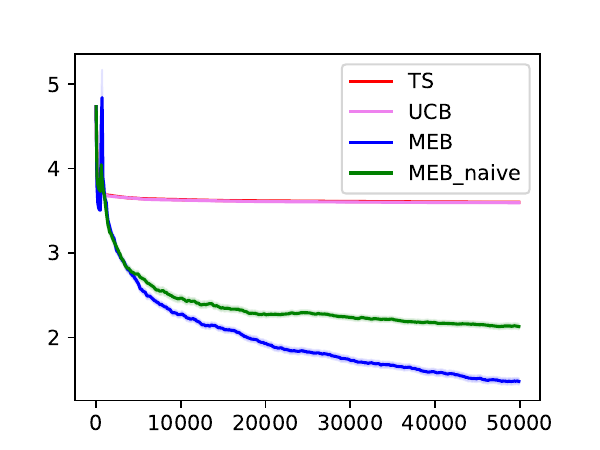}   
    \includegraphics[width = 0.3\textwidth]{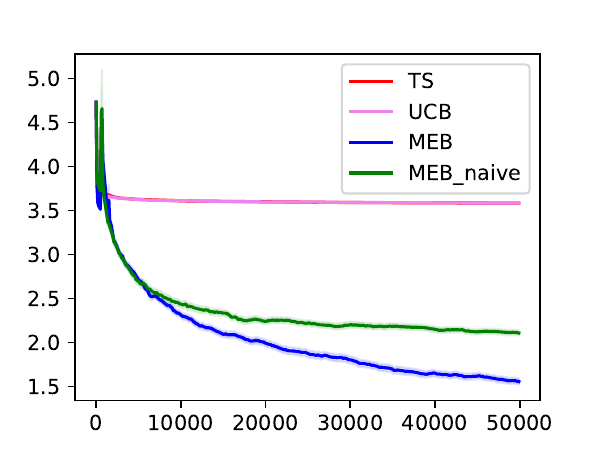}   
    \includegraphics[width = 0.3\textwidth]{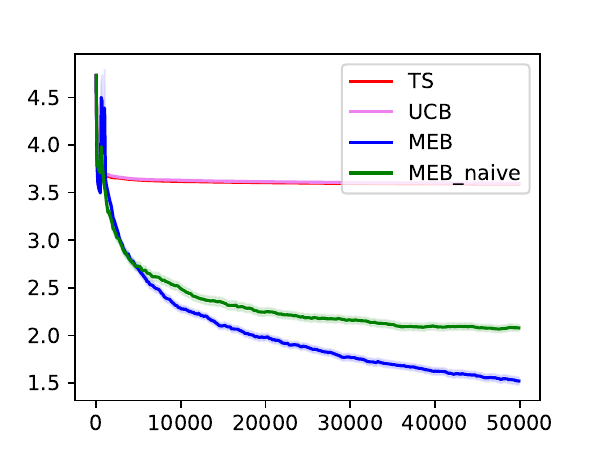}   
    \end{subfigure}
    \caption{Log-scaled L2 norm of $\hat\btheta_1 - \btheta_1^*$ of four algorithms in the synthetic environment over 50000 steps under $\sigma_{\epsilon}^2 \in \{0.1, 1.0, 2.0\}$ and $\sigma_{\eta}^2 \in \{0.01, 0.1, 1.0\}$.}
    \label{fig:synthetic_error}
\end{figure}

\begin{figure}[H]
     \begin{flushleft}
    {\quad \quad  \tiny\quad $\sigma_{\eta}^2 = 0.05 $} \hspace{35mm} {\tiny\quad $\sigma_{\eta}^2 = 0.1 $} \hspace{35mm} {\tiny\quad $\sigma_{\eta}^2 = 5.0 $}
    \end{flushleft}
    \centering
    \begin{subfigure}{1\textwidth}
    \rotatebox{90}{\tiny\quad $\sigma_{\epsilon}^2 = 0.1$}
    \includegraphics[width = 0.3\textwidth]{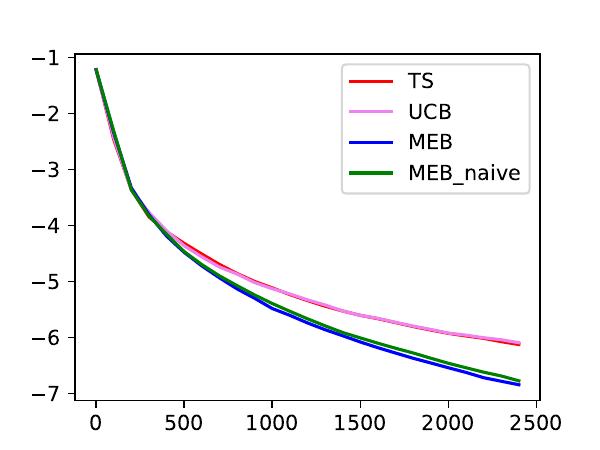}   
    \includegraphics[width = 0.3\textwidth]{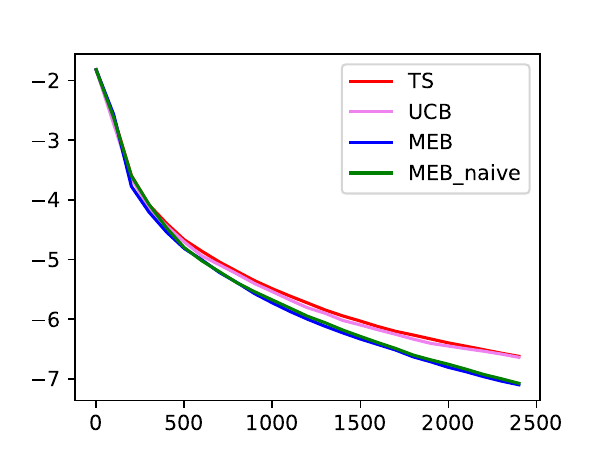}   
    \includegraphics[width = 0.3\textwidth]{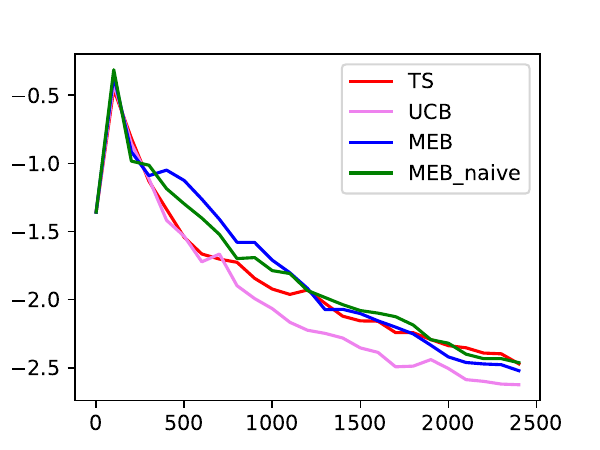}   
    \end{subfigure}
    \begin{subfigure}{1\textwidth}
    \rotatebox{90}{\tiny\quad $\sigma_{\epsilon}^2 = 1.0$}
    \includegraphics[width = 0.3\textwidth]{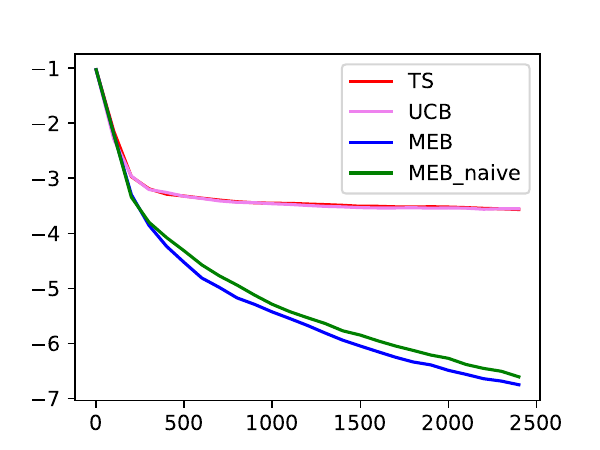}   
    \includegraphics[width = 0.3\textwidth]{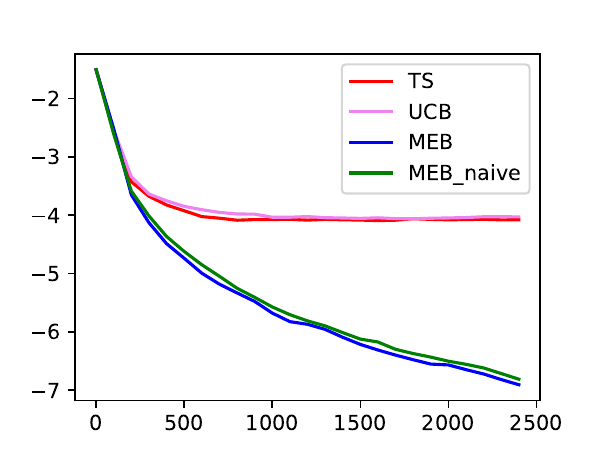}   
    \includegraphics[width = 0.3\textwidth]{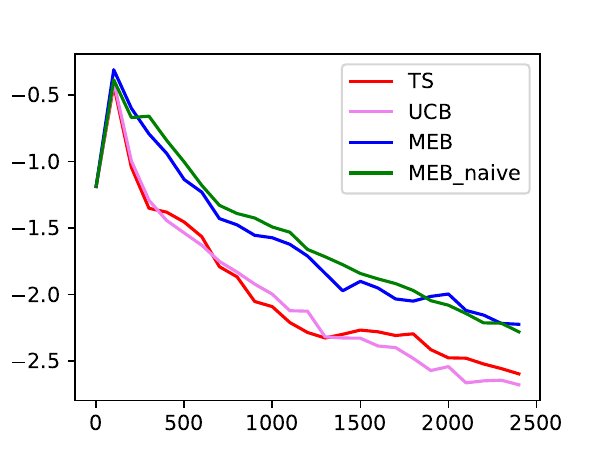}   
    \end{subfigure}
    \begin{subfigure}{1\textwidth}
    \rotatebox{90}{\tiny\quad $\sigma_{\epsilon}^2 = 2.0$}
    \includegraphics[width = 0.3\textwidth]{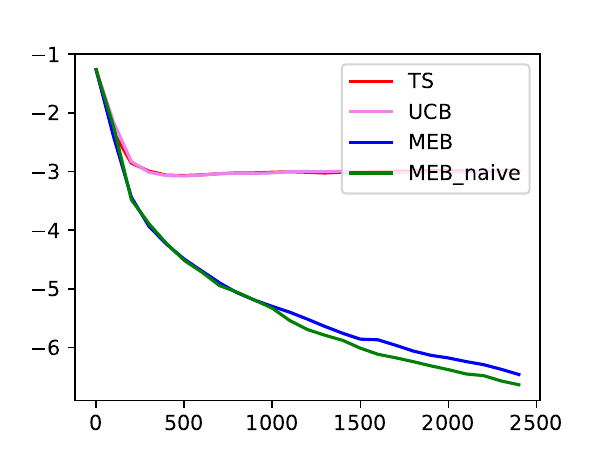}   
    \includegraphics[width = 0.3\textwidth]{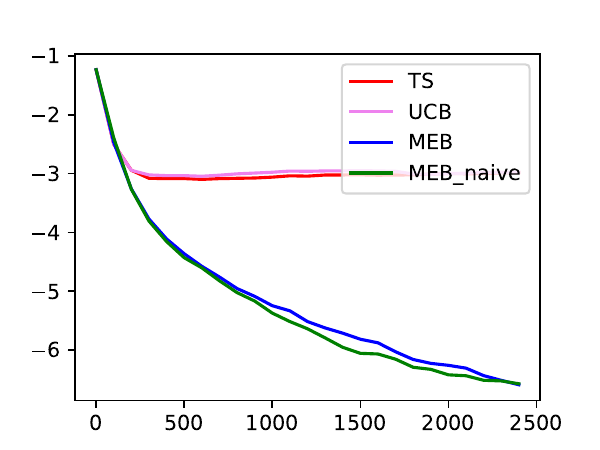}   
    \includegraphics[width = 0.3\textwidth]{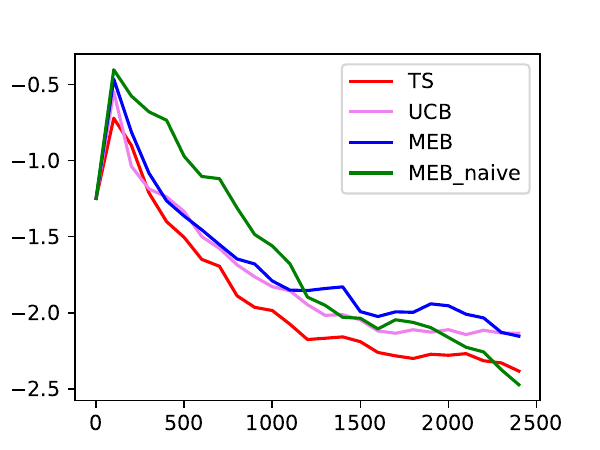}   
    \end{subfigure}
    \caption{Log-scaled L2 norm of $\hat\btheta_1 - \btheta_1^*$ of four algorithms in the real-data environment based on HeartStep V1 over 2500 steps under $\sigma_{\epsilon}^2 \in \{0.1, 1.0, 2.0\}$ and $\sigma_{\eta}^2 \in \{0.05, 0.1, 5.0\}$.}
    \label{fig:real-data_error}
\end{figure}

\subsection{Average regret with standard deviation}

\begin{table}[H]
\centering
\caption{Average regret for both synthetic environment and real-data environment under different combinations of $\sigma_{\eta}^2$ and $\sigma_{\epsilon}^2$. The numbers in parentheses are  the standard deviations calculated from 100 independent runs.}
\begin{subtable}[h]{1\textwidth}
\caption{Average regret in the synthetic environment over $50000$ steps with clipping probability $p = 0.2$. }
\centering
\begin{tabular}{ll|lllll}
$\sigma_{\eta}^2$ & $\sigma_{\epsilon}^2$ & \texttt{TS} & \texttt{UCB} & \texttt{MEB} & \texttt{MEB-naive} & \texttt{RobustUCB} \\
\hline
\hline
0.01 & 0.1 & 0.047 (0.0015) & 0.046 (0.0015) & 0.027 (0.0011) & 0.038 (0.0013) & 0.050 (0.0051) \\
0.1  & 0.1 & 0.047 (0.0015) & 0.047 (0.0015) & 0.026 (0.0011) & 0.039 (0.0013) & 0.049 (0.0048) \\
1.0  & 0.1 & 0.048 (0.0015) & 0.048 (0.0015) & 0.027 (0.0011) & 0.038 (0.0013) & 0.044 (0.0047)\\
0.01 & 1.0 & 0.757 (0.0164) & 0.647 (0.0145) & 0.198 (0.0079) & 0.371 (0.0107) & 0.652 (0.0050)\\
0.1  & 1.0 & 0.769 (0.0160) & 0.721 (0.0156) & 0.205 (0.0080) & 0.392 (0.0110) & 0.753 (0.0056)\\
1.0  & 1.0 & 0.714 (0.0155) & 0.697 (0.0150) & 0.218 (0.0083) & 0.404 (0.0112) & 0.589 (0.0047)\\
0.01 & 2.0 & 1.492 (0.0281) & 1.504 (0.0283) & 0.358 (0.0129) & 0.616 (0.0169) & 1.608 (0.0102)\\
0.1  & 2.0 & 1.195 (0.0244) & 1.333 (0.0260) & 0.368 (0.0131) & 0.584 (0.0164) & 1.064 (0.0079)\\
1.0  & 2.0 & 1.299 (0.0257) & 1.476 (0.0277) & 0.416 (0.0139) & 0.625 (0.0170) & 1.881 (0.0114)\\
\hline
\hline
\end{tabular}
\end{subtable}

\vspace{3mm}

\begin{subtable}[h]{1\textwidth}
\centering
\caption{Average regret in the real-data environment over $2500$ steps with clipping probability $p = 0.2$. }
\begin{tabular}{ll|lllll}
$\sigma_{\eta}^2$ & $\sigma_{\epsilon}^2$ & \texttt{TS} & \texttt{UCB} & \texttt{MEB} & \texttt{MEB-naive} & \texttt{RobustUCB} \\
\hline
\hline
0.05 & 0.1 & 0.027 (0.0067) & 0.027 (0.0070) & 0.022 (0.0057) & 0.024 (0.0058) & 0.025 (0.0079) \\
0.1  & 0.1 & 0.026 (0.0057) & 0.024 (0.0053) & 0.020 (0.0046) & 0.020 (0.0046) & 0.028 (0.0079) \\
5.0  & 0.1 & 1.030 (0.0287) & 0.743 (0.0262) & 0.831 (0.0267) & 1.173 (0.0343) & 1.400 (0.0447) \\
0.05 & 1.0 & 0.412 (0.0355) & 0.408 (0.0351) & 0.117 (0.0148) & 0.112 (0.0143) & 0.226 (0.0020) \\
0.1  & 1.0 & 0.309 (0.0293) & 0.316 (0.0299) & 0.085 (0.0112) & 0.087 (0.0116) & 0.206 (0.0125) \\
5.0  & 1.0 & 1.321 (0.0417) & 0.918 (0.0304) & 1.458 (0.0422) & 1.322 (0.0388) & 1.065 (0.0400)\\
0.05 & 2.0 & 0.660 (0.0343) & 0.634 (0.0322) & 0.144 (0.0129) & 0.148 (0.0133) & 0.304 (0.0241)\\
0.1  & 2.0 & 0.740 (0.0505) & 0.704 (0.0489) & 0.151 (0.0145) & 0.155 (0.0149) & 0.432 (0.0386)\\
5.0  & 2.0 & 1.585 (0.0454) & 2.415 (0.0816) & 1.577 (0.0508) & 1.436 (0.0462) & 1.345 (0.0423)
\end{tabular}
\end{subtable}
\label{tab:synthetic_full}
\end{table}

\section{Additional explanations on the regularized least-squares (RLS) estimator the naive estimator (\ref{eq:naive-estimator}) under noisy context}\label{appendix-failure-of-naive-estimator}

\subsection{Inconsistency of the RLS estimator}

\noindent\textbf{Measurement error model and attenuation.} As briefly mentioned in the main text, a measurement error model is a regression model designed to accommodate inaccuracies in the measurement of regressors. Suppose that there is no action (i.e. set $a_\tau\equiv 0$), and $(\xb_\tau, \bepsilon_\tau, \eta_\tau)_{\tau\in[t]}$ are i.i.d., then the measurement error model is a useful tool to learn $\btheta^*_0$ from $\cH_t$ collected as follows:
\begin{equation}\label{eq::appendixB-measurement-error-model}
\begin{cases}
r_\tau = \langle\btheta_0^*, \xb_\tau\rangle + \eta_\tau,\\
\tilde\xb_\tau = \xb_\tau + \bepsilon_\tau.
\end{cases}
\end{equation}
Here in the measurement error model's perspective, $(\tilde\xb_\tau)_{\tau\in[t]}$ are regressors `measured with error', and $(r_\tau)_{\tau\in[t]}$ are dependent variables. 

Regression attenuation, a phenomenon intrinsic to measurement error models, refers to the observation that when the predictors are subject to measurement errors, the Ordinary Least Squares (OLS) estimators of regression coefficients become biased (see, for instance, \cite{carroll1995measurement}). Specifically, in simple linear regression, the OLS estimator for the slope tends to be biased towards zero. Intuitively, this is because the measurement errors effectively `dilute' the true relationship between variables, making it appear weaker than it actually is. 

\noindent\textbf{Inconsistency of the RLS estimator.} Before presenting a concrete numerical example to show the inconsistency of the RLS estimator and that it leads to bad decision-making, below we first apply the theory of measurement error model to give a heuristic argument of why the RLS estimator is inconsistent even in the simplified situation where there is no action (i.e. set $a_\tau\equiv 0$), and $(\xb_\tau, \bepsilon_\tau, \eta_\tau)_{\tau\in[t]}$ are i.i.d..

From Section 3.3.2 in \cite{carroll1995measurement}, given data $(\tilde\xb_\tau, r_\tau)_{\tau\in[t]}$ from (\ref{eq::appendixB-measurement-error-model}) with multiple covariates (in the simplified case with no action as described above), the OLS estimator for $\btheta_0^*$, denoted as $\hat\btheta_0^{OLS, t}$, consistently estimates not $\btheta_0^*$ but
\begin{equation}\label{eq::AppendixB-thetatilde}
\tilde\btheta_0^* = (\bSigma_{\xb} + \bSigma_e)^{-1}\bSigma_{\xb}\btheta_0^*,
\end{equation}
where $\bSigma_\xb = \mathrm{Var}(\xb_\tau)$, $\bSigma_e = \mathrm{Var}(\bepsilon_\tau)$.
In addition, given fixed $\lambda$, the regularized least squares (RLS) estimator
$$
\hat\btheta_0^{RLS, t}(\lambda) = \bigg(\frac{\lambda}{t}I+\frac1t\sum_{\tau\in[t]}\tilde\xb_\tau\tilde\xb_\tau^\top\bigg)^{-1}\bigg(\frac1t\sum_{\tau\in[t]}\tilde\xb_\tau r_\tau\bigg) = \bigg(\frac{\lambda}{t}I+\frac1t\sum_{\tau\in[t]}\tilde\xb_\tau\tilde\xb_\tau^\top\bigg)^{-1}\bigg(\frac1t\sum_{\tau\in[t]}\tilde\xb_\tau\tilde\xb_\tau^\top\bigg)\hat\btheta_0^{OLS, t},
$$
where $\frac1t\sum_{\tau\in[t]}\tilde\xb_\tau\tilde\xb_\tau^\top\rightarrow \mathrm{Var}(\tilde\xb_\tau) = \mathrm{Var}(\xb_\tau) + \mathrm{Var}(\bepsilon_\tau)$, and $\lambda/t\rightarrow 0$ as $t\rightarrow \infty$. This means that as $t\rightarrow \infty$, $\hat\btheta_0^{RLS, t}(\lambda)$ and $\hat\btheta_0^{OLS, t}$ converges to the same limit, which is $\tilde\btheta_0^*$. Thus, for fixed $\lambda$, as $t$ grows, $\|\hat\btheta_0^{RLS, t}(\lambda) - \btheta_0^*\|_2$ converges to $\|\tilde\btheta_0^* - \btheta_0^*\|_2$ and does not converge to zero in general.

Finally, recall that in classical bandit algorithms such as UCB, the sublinear regret relies on the key property that with high probability, for all $t$, $\|\hat\btheta_0^{RLS, t}(\lambda) - \btheta_0^*\|_{V_t} \leq \beta$, where $V_t = \lambda I + \sum_{\tau\in[t]}\tilde\xb_\tau\tilde\xb_\tau^\top$, $\beta = \tilde\cO(\sqrt{d})$. Here for a vector $\bv\in \RR^d$ and positive definite matrix $\mathbf M\in\RR^{d\times d}$, $\|\bv\|_{\mathbf M}:= \sqrt{\bv^\top \mathbf M\bv}$. We argue that this requirement generally no longer holds in the setting with measurement error $(\bepsilon_\tau)_{\tau\geq 1}$. In fact, notice that since $\tilde\xb_\tau$ is i.i.d. in this simplified setting, and $V_t = \lambda I + \sum_{\tau\in[t]}\tilde\xb_\tau\tilde\xb_\tau^\top$, we expect that $\frac1t V_t$ concentrates around $\mathrm{Var}(\tilde\xb_t)$. As long as $\lambda_{\min}(\mathrm{Var}(\tilde\xb_t))>0$, with high probability, for all $t$, $\frac1t V_t\succ c$ for some constant $c$. If this holds, 
$$
\|\hat\btheta_0^{RLS, t}(\lambda) - \btheta_0^*\|_{V_t}\geq \sqrt{ct}\|\hat\btheta_0^{RLS, t}(\lambda) - \btheta_0^*\|_2,
$$
while the last term $\|\hat\btheta_0^{RLS, t}(\lambda) - \btheta_0^*\|_2$ converges to a nonzero limit $\|\tilde\btheta_0^* - \btheta_0^*\|_2$. This indicates that in general, $\|\hat\btheta_0^{RLS, t}(\lambda) - \btheta_0^*\|_{V_t}$ scales with a rate of at least $\sqrt{t}$, and will not be uniformly bounded by $\tilde\cO(\sqrt{d})$ for all $t$.

\noindent\textbf{An example.} The following is an example where given the errors $(\bepsilon_\tau)_{\tau\in[t]}$, the RLS estimator in the classical bandit algorithms inconsistently estimates the true reward model, and in addition, it leads to bad decision-making (linear regret) in the classic bandit algorithms.

\begin{example}\label{example::UCBTSfail}
Consider the standard setting described in Section \ref{sec::setting}. Let $T = 10000$, $d = 2$, $\btheta_1^* = (1, 0)^\top$, $\btheta_0^* = (-1, 0)^\top$. Let $\xb_t$ sampled i.i.d. from $\mathrm{Unif}(\cS)$, where $\cS = \{(1, 3)^\top,\thickspace (-3, 1)^\top, \thickspace(-1, -3)^\top,\thickspace (3, -1)^\top\}$. Condition on $\xb_t$, $\bepsilon_t$ is uniformly sampled from $((\rho_0 \xb_t^{[1]}, \rho_0 \xb_t^{[1]})^\top, (-\rho_0 \xb_t^{[1]}, -\rho_0 \xb_t^{[1]})^\top)$, independent from any other variable in the history. Here $\rho_0 = 0.9$, $\xb_t^{[1]}$ denotes the first entry in $\xb_t$. We also let $\eta_t$ be i.i.d. drawn from $N(0,\thickspace 0.01)$. 

We conduct 100 independent experiments. For each experiment, we generate data according to the above, and test the performance of UCB (Algorithm 1 in \cite{chu2011contextual}) and Thompson sampling with normal priors \citep{russo2018tutorial} using noisy context $\tilde\xb_t$ instead of the true context. We choose the regularization parameter $\lambda = 1$ in the RLS estimator. Additionally, in UCB \citep{chu2011contextual}, we choose the parameter $\alpha = 1$. Figure \ref{fig::appendixB-UCBTSfail} summarizes the estimation error of the RLS estimator and the cumulative regret of both algorithms with respect to time $t$, showing both the average and standard error across the random experiments. We see that the RLS estimator is unable to estimate the true reward model well. Moreover, it is clear that the regret of both UCB and Thompson sampling is linear in the time horizon. Intuitively, this is because the direction of $\tilde \btheta_0^*$ in (\ref{eq::AppendixB-thetatilde}) is twisted compared to $\btheta_0^*$, which not only leads to inconsistent estimators, but also the optimal action altered.

\begin{figure}[h]
\centering
\includegraphics[width=1.0\textwidth]{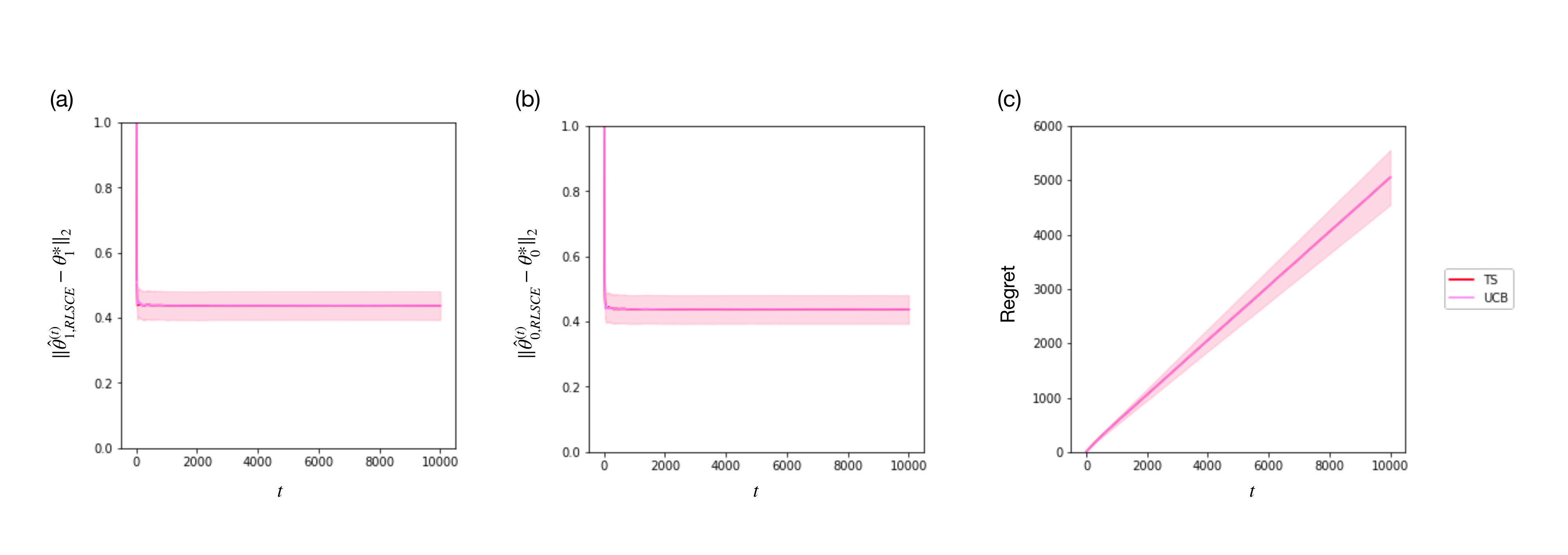}
\caption{Estimation error of the RLS estimator and cumulative regret of UCB \citep{chu2011contextual} and Thompson sampling \citep{russo2018tutorial} under contextual error in Example \ref{example::UCBTSfail}. The red and pink line corresponds to Thompson sampling and UCB respectively. The solid lines indicate the mean values, while the shaded bands represent the standard error across the independent experiments.}
\label{fig::appendixB-UCBTSfail}
\end{figure}

Finally, we note that in this setup, Assumption \ref{ass:small-error} is satisfied. This demonstrates that even if the errors $(\bepsilon_t)_{t\geq 1}$ do not affect the optimal action given $(\btheta_a^*)_{a\in\{0, 1\}}$, the poor performance of the RLS estimator may still lead to linear regret in classical bandit algorithms.
\end{example}

\subsection{Inconsistency of the naive measurement error adjusted estimator (\ref{eq:naive-estimator})}
\begin{figure}[h]
\centering
\includegraphics[width=1.0\textwidth]{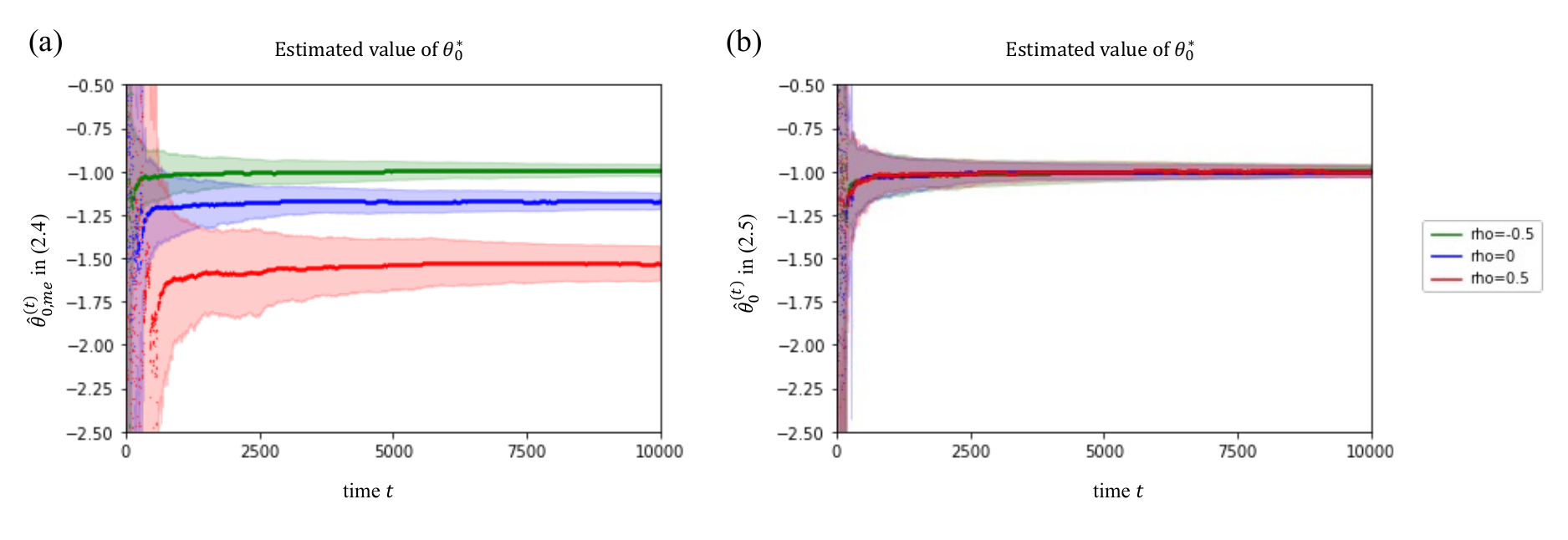}
\caption{Estimated value of $\btheta_0^*$ given the naive estimator (\ref{eq:naive-estimator}) in (a) and our proposed estimator (\ref{eq:proposed-estimator}) in (b) under different policies under 100 independent experiments. The green, blue, and red line corresponds to the policy with parameter $\rho = -0.5, 0$, and $0.5$ respectively. The solid lines indicate the mean values, while the shaded bands represent the standard deviation across the independent experiments.}
\label{fig:theta_estimation}
\end{figure}

\begin{example}\label{ex:failure-of-naive-estimator}
Let $d = 1$, $\xb_{\tau}\equiv 1$ for all ${\tau}$, and $\bepsilon_{\tau}\sim \text{Unif}(-2, 2)$ sampled independently. For the reward model, let $\btheta_0^* = -1$, $\btheta_1^*= 1$, $\eta_{\tau}\sim \text{Unif}(-0.1, 0.1)$ sampled independently. So in order to maximize expected reward, we should choose action 1 if $\xb_{\tau}$ is positive and action 0 otherwise. Suppose the agent takes the following policy that is stationary and non-adaptive to history:  
$$
\pi_{\tau}(A)=
\begin{cases}
\frac{2}{3}1_{\{A = 1\}} + \frac{1}{3}1_{\{A = 0\}},\quad\text{if }\tilde\xb_{\tau}
>\rho \\
\frac{1}{3}1_{\{A = 1\}} + \frac{2}{3}1_{\{A = 0\}},\quad\text{otherwise.}
\end{cases}
$$
Here, $\rho$ is a pre-specified constant. Figure \ref{fig:theta_estimation} (a) plots the mean and standard deviation of $\hat\btheta_{0, me}^{(t)}$ (as in (\ref{eq:naive-estimator})) given 100 independent experiments for each $t=1,\ldots, 10000$, where $\rho = -0.5, 0, 0.5$. Observe that as $t$ grows, $\hat\btheta_{0, me}^{(t)}$ 
converges to different limits for different policies. In general, 
the limit is not equal to $\btheta_0^* = -1$.

In contrast, Figure \ref{fig:theta_estimation} (b) shows the mean and standard deviation of $\hat\btheta_0^{(t)}$ (as in (\ref{eq:proposed-estimator})) given 100 independent experiments under the same setting with the same three policies as in Figure \ref{fig:theta_estimation} (a). Unlike the naive estimator (\ref{eq:naive-estimator}), the proposed estimator (\ref{eq:proposed-estimator}) quickly converges around the true value $-1$ for all three candidate policies.
\end{example}

\section{{Generalization to $K\geq 2$ actions}} \label{appendix:generalization-to-K-actions}

In this section, we assume that $\cA = \{1, 2, \ldots, K\}$ instead of $\{0, 1\}$. The standard and clipped benchmark become
\begin{equation}\label{eq:standard-benchmark-multiple-actions}
\pi_t^*(a) = 
\begin{cases}
1,\quad &\text{if }a = a_t^* = \argmax_a\langle\btheta_a^*, \xb_t\rangle\\
0,\quad &\text{otherwise, }
\end{cases}
\end{equation}
and 
\begin{equation}\label{eq:clipped-benchmark-multiple-actions}
\bar\pi_t^*(a) = 
\begin{cases}
1-(K-1)p_0,\quad &\text{if }a = a_t^*,\\
p_0,\quad &\text{otherwise. }
\end{cases}
\end{equation}

In the $K$-arm setting, we can still estimate $\btheta_a^*$ using (\ref{eq:proposed-estimator}) for each $a\in\cA$. Using the same proof ideas as Theorem \ref{thm:theta-estimation}, we get the following guarantees for estimation error (proof omitted):

\begin{theorem}\label{thm:theta-estimation-multiple-actions}
For any $t\in[T]$, let $q_t = \inf_{\tau\in[t], a\in\cA}\pi_\tau(a|\tilde\xb_\tau, \cH_{\tau-1})$. Then under Assumption \ref{ass:boundedness} and \ref{ass:min-signal}, there exist constants $C$ and $C_1$ such that as long as $q_t\geq C_1\max\left\{\frac{d(d+\log t)}{\lambda_0 t}, \frac{\xi(d+\log t)}{\lambda_0^2 t}\right\}$, with probability at least $1-\frac{4K}{t^2}$,
$$
\|\hat\btheta_a^{(t)}-\btheta_a^*\|_2\leq \frac{C(R+R_\theta)d}{\lambda_0}\max\left\{\frac{d+\log t}{q_tt}, \frac{\sqrt{\nu} + \sqrt{\xi}}{\sqrt{d}}\sqrt{\frac{d+\log t}{q_tt}}\right\}, \quad \forall a\in\cA.
$$
\end{theorem}

\begin{algorithm}[t]
	\caption{\texttt{MEB} with $K$ actions}	
	\label{alg1-multiple-actions}
	\begin{algorithmic}[1]		
\STATE \textbf{{Input}}: $(\bSigma_{e, t})_{t\in[T]}$: covariance sequence of $(\bepsilon_t)_{t\in[T]}$; $(p_0^{(t)})_{t\in[T]}$: minimum selection probability at each time $t$; $T_0$: length of pure exploration
\FOR{time $t = 1, 2, \ldots, T$}
        \IF{$t\leq T_0$}
\STATE Sample $a_t\sim \pi_t(\cdot|\tilde\xb_t, \cH_{t-1})$, where $\pi_t(a|\tilde\xb_t, \cH_{t-1}) \in[p_0^{(t)}, 1-(K-1)p_0^{(t)}]$ for all $a\in\cA$
\ELSE
\STATE Obtain updated estimators $(\hat\btheta_{a}^{(t-1)})_{a\in\cA}$ from (\ref{eq:proposed-estimator})
\STATE $\tilde a_t \leftarrow \argmax_{a\in\cA}\langle\hat\btheta_{a}^{(t-1)}, \tilde\xb_t\rangle$
\STATE Sample $a_t\sim \pi_t(\cdot|\tilde\xb_t, \cH_{t-1})$, where $\pi_t(a|\tilde\xb_t, \cH_{t-1}):=
\begin{cases}
1-(K-1)p_{0}^{(t)}, \quad\text{if }a = \tilde a_t\\
p_{0}^{(t)}, \quad\text{otherwise}
\end{cases}
$
\ENDIF
\ENDFOR
\end{algorithmic}
\end{algorithm}

\texttt{MEB} with $K$ actions is shown in Algorithm \ref{alg1-multiple-actions}. As in Theorem \ref{thm-regret}, we can control the regret of Algorithm \ref{alg1-multiple-actions} by the estimation error of (\ref{eq:proposed-estimator}). Here, Assumption \ref{ass:small-error} needs to be generalized to the following to adapt to multiple actions:

\begin{assumption}\label{ass:small-error-K-actions}
There exists a constant $\rho\in(0, 1)$ such that $\forall t\in[T]$, $\forall a_1, a_2\in\mathcal A$, $|\langle\btheta_{a_1}^* - \btheta_{a_2}^*, \bepsilon_t\rangle|\leq \rho |\langle\btheta_{a_1}^* - \btheta_{a_2}^*, \xb_t\rangle|$ almost surely.
\end{assumption}

The theorem below is a generalization of Theorem \ref{thm-regret} to multiple actions (The proof is only slightly different from Theorem \ref{thm-regret}; We briefly discuss the difference in Appendix \ref{pf:thm:regret-multiple-actions}).

\begin{theorem}\label{thm:regret-multiple-actions}
Let Assumption \ref{ass:boundedness} and \ref{ass:small-error-K-actions} hold.
\begin{itemize}
\item[(i)] For the standard setting, Algorithm \ref{alg1-multiple-actions} outputs a policy with $\text{Regret}(T; \pi^*)$ no more than 
$$
2T_0R_{\theta}+\frac{2}{1\!-\!\rho}\cdot\!\!\!\sum\limits_{t=T_0\!+1}^T\!\!\!\!\big[(K-1)p_0^{(t)}R_\theta + \!\!\max\limits_{a\in\cA}\!\|\hat\btheta_a^{(t-1)}-\btheta_a^*\|_2\big].
$$
\item[(ii)] For the clipped policy setting, Algorithm \ref{alg1-multiple-actions} with the choice of $p_0^{(t)} \equiv p_0$ outputs a policy with $\text{Regret}(T; \bar\pi^*)$ no more than 
$$
2T_0R_{\theta}+\frac{2(1\!-\!Kp_0)}{1\!-\!\rho}
\cdot\!\!\!\sum\limits_{t=T_0\!+1}^T\max\limits_{a\in\cA}\|\hat\btheta_a^{(t-1)}\!-\!\btheta_a^*\|_2.
$$
\end{itemize}
\end{theorem}

Combining Theorems \ref{thm:theta-estimation-multiple-actions} and \ref{thm:regret-multiple-actions}, we obtain the following corollary. 

\begin{corollary}\label{cor:alg1-with-theta-estimation-multiple-actions}
Let Assumption \ref{ass:boundedness}, \ref{ass:min-signal} and \ref{ass:small-error-K-actions} hold. There exist universal constants $C, C'$ such that:

(i) For the standard setting, $\forall T\geq C'\max\{(1+1/\lambda_0^{\frac94})(d+\log T)^3, (\xi/\lambda_0)^{\frac94}(d+\log T)^{\frac43}\}$, with probability at least $1\!-\!\frac{8K}{\sqrt{T}}$, Algorithm \ref{alg1-multiple-actions} with the choice of $T_0 = \lceil 2dT^{\frac23}\rceil$, $p_0^{(t)} = \min\{\frac12, t^{-\frac13}\}$ outputs a policy with $\text{Regret}(T; \pi^*)$ no more than
$$
C dT^{\frac23}\bigg\{\frac{(K-1)R_\theta}{1\!-\!\rho} + \frac{(\sqrt{\nu}+\sqrt{\xi}+1)(R\!+\!R_{\theta})}{(1\!-\!\rho)\lambda_0}\sqrt{1+\frac{\log T}{d}}\bigg\}.
$$

(ii) For the clipped policy setting, $\forall T\geq C'\max\{(d+\log T)^2/(\lambda_0p_0)^2, \xi^2/\lambda_0^4(1+\log T/d)^2\}$, with probability at least $1\!-\!\frac{8K}{\sqrt{T}}$, Algorithm \ref{alg1-multiple-actions} with the choice of $T_0 = \lceil2d\sqrt{T}\rceil$ and  $p_0^{(t)}\equiv p_0$ outputs a policy with $\text{Regret}(T; \bar\pi^*)$ no more than 
$$
CdT^{\frac12}\bigg\{R_\theta + \frac{(\sqrt{\nu}+\sqrt{\xi}+1)(1\!-\!Kp_0)(R\!+\!R_\theta)}{\sqrt{p_0}(1\!-\!\rho)\lambda_0}\sqrt{1+\frac{\log T}{d}}\bigg\}.
$$
\end{corollary}

\section{{Analysis of the proposed estimator (\ref{eq:proposed-estimator})}}\label{pf:thm:theta-estimation}

\subsection{{Proof of Theorem \ref{thm:theta-estimation}}}

We fix some $t\in[T]$, and control $\|\hat\btheta_{a}^{(t)}-\btheta_a^*\|_2$ for $a\in\{0, 1\}$. Towards this goal, we combine analysis of the two random terms $\hat\bSigma_{\tilde\xb, a}^{(t)}$ and $\hat\bSigma_{\tilde\xb, r, a}^{(t)}$ in the lemma below.
\begin{lemma}\label{lem:hat-Sigma-x-r}
Under the same assumptions of Theorem 2.1, there exists an absolute constant $C$ such that with probability at least $1-4/t^2$, both of the followings hold:
\begin{align}
\bigg\|\hat\bSigma_{\tilde\xb, a}^{(t)}-\frac{1}{t}\sum_{\tau\in[t]}\pi_{\tau}^{nd}(a)(\xb_\tau\xb_{\tau}^\top+\bSigma_{e, \tau})\bigg\|_2\leq C\max\left\{\frac{d+\log t}{q_tt}, \frac{\sqrt{\xi}}{d}\cdot \sqrt{\frac{d+\log t}{q_tt}}\right\},\label{eq:hat-Sigma-x}\\
\bigg\|\hat\bSigma_{\tilde\xb, r, a}^{(t)}-\bigg(\frac{1}{t}\sum_{\tau\in[t]}\pi_{\tau}^{nd}(a)\xb_\tau \xb_{\tau}^\top\bigg)\btheta^*_{a}\bigg\|_2\leq CR\max\left\{\frac{d+\log t}{q_tt}, \sqrt{\frac{\nu}{d}}\cdot \sqrt{\frac{d+\log t}{q_tt}}\right\}.\label{eq:hat-Sigma-x-r}
\end{align}
\end{lemma}
Proof of Lemma \ref{lem:hat-Sigma-x-r} is in Section \ref{pf:lem:hat-Sigma-x-r}.

Denote $\bDelta_1 =\hat\bSigma_{\tilde\xb, a}^{(t)}-\frac{1}{t}\sum_{\tau\in[t]}\pi_{\tau}^{nd}(a)(\xb_\tau\xb_{\tau}^\top+\bSigma_{e, \tau})$, $\bDelta_2 = \hat\bSigma_{\tilde\xb, r, a}^{(t)}-\bigg(\frac{1}{t}\sum_{\tau\in[t]}\pi_{\tau}^{nd}(a)\xb_\tau \xb_{\tau}^\top\bigg)\btheta^*_{a}$. Then
\begin{align*}
\hat\btheta_a^{(t)}&= \bigg(\hat\bSigma_{\tilde\xb, a}^{(t)} - 
\frac{1}{t}\sum_{\tau\in[t]}\pi^{nd}_\tau(a)\bSigma_{e, \tau}\bigg)^{-1}\cdot
\hat\bSigma_{\tilde\xb, r, a}^{(t)}\\
& = \bigg(\frac{1}{t}\sum_{\tau\in[t]}\pi^{nd}_\tau(a)\xb_\tau\xb_{\tau}^\top+\bDelta_1\bigg)^{-1}\cdot \left[\bigg(\frac{1}{t}\sum_{\tau\in[t]}\pi_{\tau}^{nd}(a)\xb_\tau \xb_{\tau}^\top\bigg)\btheta^*_{a}+\bDelta_2\right]\\
&= \btheta_a^* - \bJ_1 + \bJ_2,
\end{align*}
where 
\begin{align*}
\bJ_1:=\left[\frac{1}{t}\sum_{\tau\in[t]}\pi_{\tau}^{nd}(a)\xb_\tau\xb_{\tau}^\top+\bDelta_1\right]^{-1}\cdot\bDelta_1\btheta_a^*,\quad
\bJ_2:=\left[\frac{1}{t}\sum_{\tau\in[t]}\pi_{\tau}^{nd}(a)\xb_\tau\xb_{\tau}^\top+\bDelta_1\right]^{-1}\cdot\bDelta_2.
\end{align*}
It's easy to verify that under the event where both (\ref{eq:hat-Sigma-x}) and (\ref{eq:hat-Sigma-x-r}) hold, whenever 
\begin{equation}\label{eq:condition-eigv-control}
C\max\left\{\frac{d+\log t}{q_tt}, \frac{\sqrt{\xi}}{d}\cdot \sqrt{\frac{d+\log t}{q_tt}}\right\}\leq \frac{\lambda_{0}}{2d},
\end{equation}
we have 
$$
\|\bJ_1\|_2\leq \frac{2CR_{\btheta}d}{\lambda_{0}}\max\left\{\frac{d+\log t}{q_tt}, \frac{\sqrt{\xi}}{d}\cdot\sqrt{\frac{d+\log t}{q_tt}}\right\}
$$
and 
$$
\|\bJ_2\|_2\leq \frac{2CRd}{\lambda_{0}}\max\left\{\frac{d+\log t}{q_tt}, \sqrt{\frac{\nu}{d}}\cdot \sqrt{\frac{d+\log t}{q_tt}}\right\}.
$$
(\ref{eq:condition-eigv-control}) can be ensured by $t\geq C_1\max\{\frac{d(d+\log t)}{\lambda_0q_t}, \frac{\xi(d+\log t)}{\lambda_0^2 q_t}\}$, where $C_1 = \max\{2C, 4C^2\}$. Given these guarantees, we have with probability at least $1-4/t^2$,
$$
\|\hat\btheta_a^{(t)}-\btheta_a^*\|_2\leq \|\bJ_1\|_2+\|\bJ_2\|_2\leq \frac{2C(R+R_\theta)d}{\lambda_0}\max\left\{\frac{d+\log t}{q_t t}, \frac{\sqrt{\nu}+\sqrt{\xi}}{\sqrt{d}}\cdot\sqrt{\frac{d+\log t}{q_t t}}\right\}.
$$
Thus we conclude the proof.

\subsection{Additional comments on Assumption \ref{ass:min-signal}}

Following Remark \ref{remark::ass-min-signal}, all the following examples of $\{\xb_\tau\}$ allow the existence of $\lambda_{0}$ with $\pi_\tau^{nd}(a)\equiv 1/2$ given any reasonably big $t$ with high probability:
\begin{itemize}
    \item $\{\sqrt{d}\xb_\tau\}$ is an i.i.d. sequence satisfying $d\EE\xb_\tau\xb_\tau^\top\succeq \lambda_{1}I_d$, $\lambda_1>0$;
    \item $\{\sqrt{d}\xb_\tau\}$ is a weakly-dependent stationary time series (a common example is the multivariate ARMA process under regularity conditions, see e.g. \citep{banna2016bernstein}). The stationary distribution $P$ satisfies $d\EE_{\xb\sim P}\xb\xb^\top\succeq \lambda_2I_d$, $\lambda_2>0$;
    \item $\{\sqrt{d}\xb_\tau\}$ is a periodic time series such that there exists $t_0\in \mathbb N^+$ which satisfies $\frac{d}{t_0}\sum_{\tau\in(kt_0+1, (k+1)t_0]}\xb_\tau\xb_\tau^\top\succeq \lambda_3I_d$ a.s. $\forall k\in\mathbb N$.
\end{itemize}



\subsection{{Generalization to off-policy method-of-moment estimation}}

(\ref{eq:proposed-estimator}) can be generalized to a class of method-of-moment estimators for off-policy learning. In this section, we delve into the general framework of off-policy method-of-moment estimation. This framework proves valuable in scenarios where a fully parametric model class for the reward is unavailable, yet there is a desire to estimate certain model parameters using offline batched bandit data.

For simplicity, we assume that $(X_t, Y_t(a):a\in\cA)_{t\in[T]}$ are drawn i.i.d. from an unknown distribution $\cP$. At each time $t\in[T]$, the action $A_t$ is drawn from a policy $\pi_t(\cdot|X_t, \cH_{t-1})$, and the agent observes only $o_t = (X_t, A_t, Y_t(A_t))$ together with the action selection probabilities $\pi_t$. Define the history up to time $t$ as $\cH_t = \{o_\tau\}_{\tau\leq t}$. For $a_0\in\cA$, we're interested in estimating $\btheta_{a_0}^*$, a $d$-dimensional parameter in $\cP^{(a_0)}$, which is the joint distribution of $\{X_t, Y_t(a_0)\}$. 

\begin{remark}
When the context is i.i.d., the problem of estimating $\{\btheta_a^*\}_{a\in\cA}$ in Section \ref{section::estimation} is a special case of this setup by taking $\tilde\xb_t$ as $X_t$ and $r_t$ as $Y_t$.
\end{remark}

The traditional method-of-moment estimator looks for functions $f_1, \ldots, f_d$ as well as a mapping $\bphi:\RR^d\rightarrow \RR^d$, such that 
$$
\btheta_{a_0}^* = \bphi\big(\EE_{(X, Y)\sim \cP^{(a_0)}}f_1(X, Y), \ldots, \EE_{(X, Y)\sim \cP^{(a_0)}}f_d(X, Y)\big).
$$
Then, if given i.i.d. samples $(U_t, V_t)_{t\in[n]}$ from $\cP^{(a_0)}$, the estimator takes the form 
$$
\hat\btheta_{a_0} = \bphi\bigg(\frac{1}{T}\sum_{t\in[T]}f_1(U_t, V_t), \ldots, \frac{1}{T}\sum_{t\in[T]}f_d(U_t, V_t)\bigg).
$$
In fact, the naive estimator (\ref{eq:naive-estimator}) is of this form. It is clear that we cannot use this estimator for offline batched data $\cH_T$: There are no i.i.d. samples from $\cP^{(a_0)}$ because of the policy $\{\pi_t\}_{t\in[T]}$. Instead, we propose the following estimator: 
$$
\hat\btheta_{a_0} = \bphi\bigg(\frac{1}{T}\sum_{t\in[T]}W_tf_1(X_t, Y_t), \ldots, \frac{1}{T}\sum_{t\in[T]}W_tf_d(X_t, Y_t)\bigg),
$$
where $W_t = 1_{\{A_t = a_0\}}\frac{\pi^{nd}(A_t)}{\pi_t(A_t|X_t, \cH_{t-1})}$ for a data-independent probability distribution $\pi^{nd}$ on $\cA$. Similar to the proof of Theorem \ref{thm:theta-estimation}, it's not difficult to see that $\hat\btheta_{a_0}$ is consistent under mild conditions. In fact, (\ref{eq:proposed-estimator}) is a special case of this estimator when $\pi_{\tau}^{nd}$ does not depend on $\tau$. A more detailed analysis is left for future work.

\section{{Analysis of \texttt{MEB}}}\label{pf:cor:alg1-with-theta-estimation}

\subsection{{Additional comments on Example \ref{ex::unknown-xt-linear-regret}}}

In example \ref{ex::unknown-xt-linear-regret}, the joint distribution of $(\xb_t, \tilde\xb_t)$ is: 
\begin{align*}
&\PP((\xb_t, \tilde\xb_t) = (0.2, 1)) = 0.3,\quad \PP((\xb_t, \tilde\xb_t) = (0.2, -1)) = 0.2,\\
&\PP((\xb_t, \tilde\xb_t) = (-0.2, 1)) = 0.2,\quad \PP((\xb_t, \tilde\xb_t) = (-0.2, -1)) = 0.3.
\end{align*}
The optimal action given $(\btheta_a^*)_{a\in\{0, 1\}}$ and $\xb_t$ is 
$$
a_t^*(\xb_t) = \argmax_a\langle\btheta_a^*, \xb_t\rangle = 
\begin{cases}
1,\quad& \text{if }\xb_t = 0.2,\\
0,\quad& \text{if }\xb_t = -0.2.
\end{cases}
$$
In the standard bandit setting, the benchmark policy is 
$$
\pi_t^*(a) = 
\begin{cases}
1,\quad& \text{if }a = a_t^*(\xb_t),\\
0,\quad& \text{otherwise}.
\end{cases}
$$
For any policy $\pi = (\pi_t)_{t\geq 1}$, the instantaneous regret at time $t$ is 
$$
\textrm{Regret}_t(\pi_t, \pi_t^*) = 0.4\EE_{a_t\sim \pi_t(\cdot|\tilde\xb_t, \cH_{t-1})}1_{\{a_t\neq a_t^*(\xb_t)\}}.
$$
Even if both $(\btheta_a^*)_{a\in\{0, 1\}}$ and the joint distribution of $(\xb_t, \tilde\xb_t)$ are known, $\pi_t$ can only depend on $\tilde\xb_t$ and history, and cannot be based on $\xb_t$. Thus, there is always a (constant) positive probability that the action $a_t$ sampled from $\pi_t$ does not match $a_t^*(\xb_t)$ (otherwise, $a_t$ sampled from $\pi_t(\cdot|\tilde\xb_t, \cH_{t-1})$ should be equal to $a_t^*(\xb_t)$ a.s.). Thus, the standard cumulative regret will be linear in the time horizon. 


\subsection{Proof of Theorem \ref{thm-regret}}\label{pf:thm-regret}

We first prove the lemma below. Its proof is in Appendix \ref{pf::lem::at-star=at-dagger}.
\begin{lemma}\label{lem::at-star=at-dagger}
Under Assumption \ref{ass:small-error}, we have $a_t^* = a_t^\dagger:= \argmax_a\langle\btheta_a^*, \tilde\xb_t\rangle$. Consequently, $\pi_t^* = \pi_t^\dagger$, $\bar\pi_t^* = \bar\pi_t^\dagger$ (given a fixed minimum action selection probability $p_0$), where 
$$
\pi_t^\dagger(a) = 
\begin{cases}
1,\quad& \text{if }a = a_t^\dagger,\\
0,\quad& \text{otherwise},
\end{cases}
\quad
\bar\pi_t^\dagger(a) = 
\begin{cases}
1-p_0,\quad& \text{if }a = a_t^\dagger,\\
p_0,\quad& \text{otherwise}.
\end{cases}
$$
\end{lemma}

In the below, we define 
\begin{align*}
\widehat{\text{Regret}}_t(\pi, \pi^*) : = \EE_{a\sim \pi_t^*}\langle \btheta^*_{a}, \tilde\xb_t\rangle - \EE_{a\sim \pi_t}\langle \btheta^*_{a}, \tilde\xb_t\rangle, \\
\widehat{\text{Regret}}_t(\pi, \bar\pi^*) : = \EE_{a\sim \bar\pi_t^*}\langle \btheta^*_{a}, \tilde\xb_t\rangle - \EE_{a\sim \pi_t}\langle \btheta^*_{a}, \tilde\xb_t\rangle.
\end{align*}

\noindent\textbf{Standard setting.} In the standard setting, we give the lemma below (proof in Appendix \ref{pf::lem::standard-setting-regret-hat}).
\begin{lemma}\label{lem::standard-setting-regret-hat}
Under the assumptions of Theorem \ref{thm-regret}, at any time $t>T_0$,
$$
\widehat{\text{Regret}}_t(\pi, \pi^*)\leq 2p_0^{(t)}R_\theta + 2\max_{a\in\{0, 1\}}\|\hat\btheta_a^{(t-1)} - \btheta_a^*\|_2.
$$
\end{lemma}

Note that for any time $t>T_0$, the instantaneous regret at time $t$: $\text{Regret}_t(\pi, \pi^*) = (\pi_t^*(1)-\pi_t(1))\langle\btheta_1^* - \btheta_0^*, \xb_t\rangle$, and that $\widehat{\text{Regret}}_t(\pi, \pi^*) = (\pi_t^*(1)-\pi_t(1))\langle\btheta_1^* - \btheta_0^*, \tilde\xb_t\rangle$. Moreover,
\begin{align*}
|\langle\btheta_1^* - \btheta_0^*, \tilde\xb_t\rangle|
= &|\langle\btheta_1^* - \btheta_0^*, \xb_t\rangle + \langle\btheta_1^* - \btheta_0^*, \bepsilon_t\rangle|\\
\geq &|\langle\btheta_1^* - \btheta_0^*, \xb_t\rangle| - |\langle\btheta_1^* - \btheta_0^*, \bepsilon_t\rangle|\\
\geq & \frac1\rho |\langle\btheta_1^* - \btheta_0^*, \bepsilon_t\rangle| - |\langle\btheta_1^* - \btheta_0^*, \bepsilon_t\rangle|\\
= & \frac{1-\rho}{\rho} |\langle\btheta_1^* - \btheta_0^*, \bepsilon_t\rangle|.
\end{align*}
Here we used Assumption \ref{ass:small-error}. Thus we have
\begin{align*}
\text{Regret}_t(\pi, \pi^*)
& = \widehat{\text{Regret}}_t(\pi, \pi^*) - (\pi_t^*(1)-\pi_t(1))\langle\btheta_1^* - \btheta_0^*, \bepsilon_t\rangle\\
&\leq \widehat{\text{Regret}}_t(\pi, \pi^*) + |\pi_t^*(1)-\pi_t(1)|\cdot \frac{\rho}{1-\rho}|\langle\btheta_1^* - \btheta_0^*, \tilde\xb_t\rangle|\\
& = \frac{1}{1-\rho}\widehat{\text{Regret}}_t(\pi, \pi^*).
\end{align*}
Combining Lemma \ref{lem::standard-setting-regret-hat}, we obtain that for any $t>T_0$, 
$$
\text{Regret}_t(\pi, \pi^*)\leq \frac2{1-\rho}\big(p_0^{(t)}R_\theta + \max_{a\in\{0, 1\}}\|\hat\btheta_a^{(t-1)} - \btheta_a^*\|_2\big).
$$

Finally, when $t\leq T_0$, since $\|\xb_t\|\leq 1$, $\|\btheta_a^*\|\leq R_\theta$, the instantaneous regret $\text{Regret}_t(\pi, \pi^*)\leq 2R_\theta$. We conclude the proof by summing up all the instantaneous regret terms.

\noindent\textbf{Clipped policy setting.} In the clipped policy setting, we give the lemma below (proof in Appendix \ref{pf::lem::clipped-setting-regret-hat}).

\begin{lemma}\label{lem::clipped-setting-regret-hat}
Under the assumptions of Theorem \ref{thm-regret}, at any time $t>T_0$, 
$$
\widehat{\text{Regret}}_t(\pi, \bar\pi^*)
\leq 2(1-2p_0) \max_{a\in\{0, 1\}}\|\hat\btheta_a^{(t-1)} - \btheta_a^*\|_2.
$$
\end{lemma}

Note that the instantaneous regret at time $t$: $\text{Regret}_t(\pi, \bar\pi^*) = (\bar\pi_t^*(1)-\pi_t(1))\langle\btheta_1^* - \btheta_0^*, \xb_t\rangle$, and that $\widehat{\text{Regret}}_t(\pi, \bar\pi^*) = (\bar\pi_t^*(1)-\pi_t(1))\langle\btheta_1^* - \btheta_0^*, \tilde\xb_t\rangle$. Similar to the standard setting, for $t>T_0$, under Assumption \ref{ass:small-error}, we have 
$$
\text{Regret}_t(\pi, \bar\pi^*)\leq \frac{1}{1-\rho}\widehat{\text{Regret}}_t(\pi, \bar\pi^*)\leq \frac{2(1-2p_0)}{1-\rho}\max_{a\in\{0, 1\}}\|\hat\btheta_a^{(t-1)} - \btheta_a^*\|_2.
$$
We conclude the proof by summing up all the instantaneous regret terms, and noticing that for $t\leq T_0$, $\text{Regret}_t(\pi, \bar\pi^*)\leq 2R_\theta$.

\noindent\textbf{Results with a high-probability version of Assumption \ref{ass:small-error}.} As briefly mentioned in the main paper, Assumption \ref{ass:small-error} can be weakened to the inequalities holding with high probability. Now instead of Assumption \ref{ass:small-error}, we assume the following:
\begin{assumption}\label{ass:small-error-high-probability}
There exist constants $\rho\in(0, 1)$, $c_e\in[0, 1]$ such that $\sum_{t=1}^T\PP(A_t^c)\leq c_e$. Here $A_t$ denotes the event $\{|\langle\bm{\delta}_\theta, \bepsilon_t\rangle|\leq \rho|\langle\bm{\delta}_\theta, \xb_t\rangle|\}$, and $\bm{\delta}_\theta = \btheta_1^* - \btheta_0^*$.
\end{assumption}

It's easy to see that the result of Lemma \ref{lem::at-star=at-dagger} hold at time $t$ under the event $A_t$. Further, following the same arguments, we obtain that under Assumption \ref{ass:boundedness}, the results for either the standard setting or the clipped policy setting hold under the event $\cap_{t = T_0+1}^TA_t$. Therefore, with Assumption \ref{ass:boundedness} and \ref{ass:small-error-high-probability}, in either setting, the results in Theorem \ref{thm-regret} hold with probability at least $1-c_e$.

\subsection{Proof of Corollary \ref{cor:alg1-with-theta-estimation}}\label{Appendix:subsection:Proof-of-corollary-2.1}

\noindent\textbf{Standard setting.} First, notice that $q_t = \min_{\tau\leq t, a\in\{0, 1\}}\pi_\tau(a|\tilde\xb_t, \cH_{\tau-1}) = p_0^{(t)}$, since $p_0^{(t)}$ is monotonically decreasing in $t$. Theorem \ref{thm:theta-estimation} indicates that, as long as $\forall t> T_0$,
\begin{equation}\label{eq:standard-benchmark-condition1}
p_0^{(t)}\geq {C_1}\max\left\{\frac{d(d+\log t)}{\lambda_0 t}, \frac{\xi(d+\log t)}{\lambda_0^2t}\right\},
\end{equation}
then with probability at least $1-\frac{8}{t^2}$, $\forall a\in\{0, 1\}$,
\begin{align}
\|\hat\btheta_a^{(t)}-\btheta_a^*\|_2
&\leq \frac{C(R+\!R_\theta)d}{\lambda_0}\max\left\{\!\frac{d+\log t}{q_t t}, \frac{\sqrt{\nu} + \sqrt{\xi}}{\sqrt{d}}\sqrt{\frac{d+\log t}{q_t t}}\right\}\label{eq:standard-benchmark-estimation-error}
\end{align}

Plug it into Theorem \ref{thm-regret}, we have that with high probability, 
\begin{align*}
\text{Regret}(T;\pi^*)
&\leq 2R_\theta\cdot \lceil 2dT^{2/3}\rceil + \frac2{1-\rho} I_1,
\end{align*}
where
\begin{align*}
I_1& = \sum_{t=T_0+1}^T
\bigg(p_0^{(t)}R_\theta + \max_a\|\hat \btheta_a^{(t-1)} - \btheta^*_a\|_2\bigg)\\
&\leq \sum_{t=T_0+1}^T t^{-\frac13}R_\theta + \sum_{t=T_0}^{T-1}\frac{C(R+\!R_\theta)d}{\lambda_0}\max\left\{\!\frac{d+\log t}{t^{\frac23}}, \frac{\sqrt{\nu} + \sqrt{\xi}}{\sqrt{d}}\sqrt{\frac{d+\log t}{t^{\frac23}}}\right\}\\
&\leq 2R_\theta T^{\frac23} + \frac{C(R+\!R_\theta)d}{\lambda_0}\cdot \sum_{t=T_0}^{T-1} \frac{d+\log t}{t^{\frac23}} + \frac{C(R+\!R_\theta)d}{\lambda_0}\cdot\sum_{t=T_0}^{T-1}\frac{\sqrt{\nu} + \sqrt{\xi}}{\sqrt{d}}\sqrt{\frac{d+\log t}{t^{\frac23}}}\\
&\leq 2R_\theta T^{\frac23} + \frac{3C(R+\!R_\theta)}{\lambda_0}d(d+\log T)T^{\frac13} + \frac{3C(\sqrt{\nu} + \sqrt{\xi})(R+\!R_\theta)}{2\lambda_0}\sqrt{d(d+\log T)}T^{\frac23}\\
&\leq 2R_\theta T^{\frac23} +\frac{C'}{\lambda_0}(\sqrt{\nu} + \sqrt{\xi} + 1)(R+\!R_\theta)\sqrt{d(d+\log T)} T^{\frac23},
\end{align*}
for a universal constant $C'$, where the last inequality holds if in addition, $T\geq[d(d+\log T)]^{\frac32}$.

The proof is concluded by combining the above requirement for $T$ as well as \eqref{eq:standard-benchmark-condition1}.

\noindent\textbf{Clipped policy setting.} Similar to the standard setting, according to Theorem \ref{thm:theta-estimation}, as long as $\forall t> T_0$,
\begin{equation}\label{eq:clipped-benchmark-condition1}
p_0\geq {C_1}\max\left\{\frac{d(d+\log t)}{\lambda_0 t}, \frac{\xi(d+\log t)}{\lambda_0^2t}\right\},
\end{equation}
then with probability at least $1-\frac{8}{t^2}$, $\forall a\in\{0, 1\}$,
\begin{align}
\|\hat\btheta_a^{(t)}-\btheta_a^*\|_2
&\leq \frac{C(R+\!R_\theta)d}{\lambda_0}\max\left\{\!\frac{d+\log t}{p_0 t}, \frac{\sqrt{\nu} + \sqrt{\xi}}{\sqrt{d}}\sqrt{\frac{d+\log t}{p_0 t}}\right\}\label{eq:clipped-benchmark-estimation-error}
\end{align}

Plug it into Theorem \ref{thm-regret}, we have that with high probability, 
\begin{align*}
\text{Regret}(T;\bar \pi^*)
&\leq 2R_\theta\cdot \lceil 2dT^{1/2}\rceil + \frac{2(1-2p_0)}{1-\rho} I_2,
\end{align*}
where
\begin{align*}
I_2& = \sum_{t=T_0+1}^T \max_a\|\hat\btheta_a^{(t-1)}-\btheta^*_a\|_2\\
&\leq \sum_{t=T_0}^{T-1}\frac{C(R+\!R_\theta)d}{\lambda_0}\max\left\{\!\frac{d+\log t}{p_0 t}, \frac{\sqrt{\nu} + \sqrt{\xi}}{\sqrt{d}}\sqrt{\frac{d+\log t}{p_0 t}}\right\}\\
&\leq \frac{C(R+\!R_\theta)d}{\lambda_0}\sum_{t=T_0}^{T-1}\frac{d+\log t}{p_0 t} + \frac{C(R+\!R_\theta)d}{\lambda_0}\sum_{t=T_0}^{T-1}\frac{\sqrt{\nu} + \sqrt{\xi}}{\sqrt{d}}\sqrt{\frac{d+\log t}{p_0 t}}\\
&\leq \frac{2C(R+\!R_\theta)}{\lambda_0}d(d+\log T)\log T + \frac{2C(R+\!R_\theta)}{\lambda_0}\frac{\sqrt{\nu} + \sqrt{\xi}}{\sqrt{p_0}}\sqrt{d(d+\log T)}\sqrt{T}\\
&\leq \frac{2C(R+R_\theta)(\sqrt{\nu}+\sqrt{\xi}+1)}{\lambda_0\sqrt{p_0}}\sqrt{d(d+\log T)}\sqrt{T},
\end{align*}
where the last inequality holds if in addition, $T\geq d(d+\log T)\log^2 T$. 

The proof is concluded by plugging the above into the regret upper bound formula and combining the requirements for $T$.

\noindent\textbf{Results with a high-probability version of Assumption \ref{ass:small-error}.} Recall that Assumption \ref{ass:small-error-high-probability} is a weakened version of Assumption \ref{ass:small-error} with a high-probability statement. Given Assumptions \ref{ass:boundedness}, \ref{ass:min-signal}, and \ref{ass:small-error-high-probability} (instead of \ref{ass:small-error}), in the standard setting, the results of Theorem \ref{cor:alg1-with-theta-estimation} hold as long as (\ref{eq:standard-benchmark-estimation-error}) or (\ref{eq:clipped-benchmark-estimation-error}) for all $t>T_0$, $a\in\{0, 1\}$ and under the event $\cap_{t = T_0+1}^TA_t$. Thus, we deduce that the regret upper bound in (i) hold with probability at least $1-16/\sqrt{T} - c_e$. Similarly, given Assumptions \ref{ass:boundedness}, \ref{ass:min-signal}, and \ref{ass:small-error-high-probability}, in the clipped benchmark setting, the regret upper bound in (ii) hold with probability at least $1-16/\sqrt{T} - c_e$.

\subsection{\texttt{MEB} with infrequent model update}
\begin{algorithm}[t]
\caption{\texttt{MEB} with infrequent model update}	
\label{alg2}
\begin{algorithmic}		
\STATE \textbf{{Input}}: $(\bSigma_{e, t})_{t\in[T]}$: covariance sequence of $(\bepsilon_t)_t$; $(p_0^{(t)})_{t\in[T]}$: minimum selection probability at each time $t$; $\cS\subset [T]$: set of time points to update model estimates
\STATE $\hat\btheta_a \leftarrow 0$, $\forall a\in\{0, 1\}$  \thickspace \thickspace {\color{blue}$\%$\thinspace$\hat\btheta_a$ stores the most recent updated estimate of $\btheta_a^*$, only update if $t\in\cS$}
        \FOR{time $t = 1, 2, \ldots, T$}
        \IF{$\min_{s\in\cS}s\geq t$}
        \STATE Sample $a_t\sim \pi_t(\cdot|\tilde\xb_t, \cH_{t-1})$, where $\pi_t(a|\tilde\xb_t, \cH_{t-1}) \in[p_0^{(t)}, 1-p_0^{(t)}]$ for all $a\in\{0, 1\}$\\
        {\color{blue} $\%$ If the model has never been learned before, explore with equal probability}
        \STATE \textbf{continue}
        \ENDIF
        \STATE $\tilde a_t \leftarrow \argmax_{a\in\{0, 1\}}\langle\hat\btheta_{a}, \tilde\xb_t\rangle$
        \STATE Sample $a_t\sim \pi_t(\cdot|\tilde\xb_t, \cH_{t-1})$, where $\pi_t(a|\tilde\xb_t, \cH_{t-1}):=
        \begin{cases}
        1-p_{0}^{(t)}, \quad\text{if }a = \tilde a_t\\
        p_{0}^{(t)}, \quad\text{otherwise}
        \end{cases}
        $
        \IF{$t\in\cS$}
        \STATE $\hat\btheta_a\leftarrow \hat\btheta_a^{(t)}$ as in (\ref{eq:proposed-estimator})
        \ENDIF
        \ENDFOR
	\end{algorithmic}
\end{algorithm}

As mentioned at the end of Section \ref{section:meb}, in certain scenarios (e.g. when $d$ is large), we can save computational resources by updating the estimates of $(\btheta_a^*)_{a\in\{0, 1\}}$ less frequently. In Algorithm \ref{alg2}, we propose a variant of Algorithm \ref{alg1}. At each time $t$, given the noisy context $\tilde\xb_t$, the algorithm computes the best action $\tilde a_t$ according to the most recently updated estimators of $(\btheta_a^*)_{a\in\{0, 1\}}$. Then, it samples $\tilde a_t$ with probability $1-p_0^{(t)}$ and keeps an exploration probability of $p_0^{(t)}$ to sample the other action. In the meantime, the agent only has to update the estimate of $(\btheta_a^*)_{a\in\{0, 1\}}$ once in a while to save computation power: The algorithm specifies a subset $\cS\subset[T]$ and updates the estimators according to (\ref{eq:proposed-estimator}) only when $t\in\cS$. 

Under mild conditions, Algorithm \ref{alg2} achieves the same order of regret upper bound as Algorithm \ref{alg1}, as seen from Theorem \ref{thm-regret-S} and Corollary \ref{cor:alg1-with-theta-estimation-S} below. They are modified versions of Theorem \ref{thm-regret} and Corollary \ref{cor:alg1-with-theta-estimation}.

\begin{theorem}\label{thm-regret-S}
Let $s_{\min}:=\min_{s\in\cS}s$ be the first time Algorithm \ref{alg2} updates the model. Suppose Assumption \ref{ass:boundedness} and \ref{ass:small-error} hold. 

\begin{itemize}
\item[(i)] For the standard setting, for any $T_0\geq s_{\min}$, Algorithm \ref{alg2} outputs a policy such that 
$$
\text{Regret}(T; \pi^*)\leq 2T_0R_{\theta}+\frac{2}{1-\rho}\sum_{t\in(T_0, T]}\big(p_0^{(t)}R_\theta + \max_{a\in\{0, 1\}}\|\hat\btheta_a^{(s_t)}-\btheta_a^*\|_2\big).
$$
\item[(ii)] For the clipped policy setting, for any  $T_0\geq s_{\min}$, Algorithm \ref{alg2} with the choice of $p_0^{(t)}\equiv p_0$ outputs a policy such that
$$
\text{Regret}(T; \bar\pi^*)\leq 2T_0R_{\theta}+\frac{2(1-2p_0)}{1-\rho}\sum\nolimits_{t\in(T_0, T]}\max_{a\in\{0, 1\}}\|\hat\btheta_a^{(s_t)}-\btheta_a^*\|_2.
$$
\end{itemize}
Here for any $t\in[T]$, $s_t := \max\{s\in\cS: s<t\}$.
\end{theorem}

The proof of Theorem \ref{thm-regret-S} is very similar to that of Theorem \ref{thm-regret}, and is thus omitted.

\begin{corollary}\label{cor:alg1-with-theta-estimation-S}
Let Assumption \ref{ass:boundedness} to \ref{ass:small-error} hold. There exist constants $C, C'$ such that:
\begin{itemize}
    \item[(i)] In the standard setting, as long as the set of model update times $\cS$ satisfies: (a) $s_{\min}\leq dT^{2/3}$; (b) $\forall t\in(dT^{2/3}, T]$, $s_t = \max\{s\in\cS: s<t\} \geq \alpha t$ for some constant $\alpha\in(e^{-d}, 1)$, then $\forall T\geq C'/\alpha\cdot \max\left\{(1+1/\lambda_0^{\frac94})(d+\log T)^3, (\xi/\lambda_0)^{\frac94}(d+\log T)^{\frac43}\right\}$, with probability at least $1-\frac{16}{\sqrt{T}}$, Algorithm \ref{alg2} with the choice of $p_0^{(t)} = \min\{\frac12, t^{-1/3}\}$ achieves $\text{Regret}(T; \pi^*)\leq CdT^{\frac23}\left\{\frac{R_\theta}{\alpha}+\frac{R_\theta}{1-\rho}+\frac{(\sqrt{\nu}+\sqrt{\xi}+1)(R+R_\theta)}{\alpha^{1/3}\lambda_{0}(1-\rho)}\sqrt{1+\frac{\log T}{d}}\right\}$.
    \item[(ii)] In the clipped policy setting, as long as the set of model update times $\cS$ satisfies: (a) $s_{\min}\leq d\sqrt{T}$; (b) $\forall t\in(d\sqrt{T}, T]$, $s_t = \max\{s\in\cS: s<t\} \geq \alpha t$ for some constant $\alpha\in(e^{-d}, 1)$, then for any $T$ s.t. $T\geq C'/\alpha\cdot \max\{(d+\log T)^2/(\lambda_0p_0)^2, \xi^2/\lambda_0^4(1+\log T/d)^2\}$, with probability at least $1-\frac{16}{\sqrt{T}}$, Algorithm \ref{alg2} with the choice of $p_0^{(t)}\equiv p_0$ achieves: $\text{Regret}(T; \bar\pi^*)\leq Cd\sqrt{T}\left\{\frac{R_\theta}{\alpha}+\frac{(\sqrt{\nu}+\sqrt{\xi} + 1)(1-2p_0)(R+R_\theta)}{\lambda_{0}(1-\rho)\sqrt{\alpha p_0}}\sqrt{1+\frac{\log T}{d}}\right\}$.
\end{itemize}
\end{corollary}

The proof of Corollary \ref{cor:alg1-with-theta-estimation-S} can be directly obtained by combining Theorem \ref{thm:theta-estimation} and \ref{thm-regret-S} with $T_0 = dT^{2/3}/\alpha$ in the standard setting and $T_0 = dT^{1/2}/\alpha$ in the clipped benchmark setting. Thus, the proof is omitted here.

In Corollary \ref{cor:alg1-with-theta-estimation-S}, condition (a) and (b) essentially requires Algorithm \ref{alg2} not to start learning the model too late, and to keep updating the learned model at least at time points with a `geometric' growth rate. This covers a wide range of choices of $\cS$ in practice. Two typical examples of $\cS$ could be: (1) $\cS=\{t\in[T]: t=kt_0\text{ for some }k\in\mathbb N^+\}$ (the model is learned every $t_0$ time points routinely, where $t_0$ is a constant integer); (2) If $1/\alpha\in \mathbb N^+$, $\cS=\{t\in[T]: t=(1/\alpha)^k\text{ for some }k\in\mathbb N^+\}$ (the model only needs to be learned $\cO(\log T)$ times to save computation).

\section{{Analysis with estimated error variance}}\label{appendix:estimated-error-variance}

We consider the setting where at each time $t$, the agent does not have access to $\bSigma_{e, t}$, but has a (potentially adaptive) estimator $\hat\bSigma_{e, t}$. In this setting, we estimate the model using (\ref{eq::proposed-estimator-estimated-Sigma}) instead of (\ref{eq:proposed-estimator}) and plug into Algorithm \ref{alg1}. The following theorem controls the estimation error of $\tilde\btheta_a^{(t)}$. Note that compared to Theorem \ref{thm:theta-estimation}, the additional error caused by inaccuracty of $\hat\bSigma_{e, t}$ can be controlled by $\bDelta_t(a):= \frac{1}{t}\sum_{\tau\in[t]}\pi_\tau^{nd}(a)(\hat\bSigma_{e, \tau} - \bSigma_{e, \tau})$, the weighted average of the estimation errors $(\hat\bSigma_{e, \tau} - \bSigma_{e, \tau})_{\tau\in[t]}$.

\begin{theorem}\label{thm:theta-estimation-estimated-Sigma}
Recall that $q_t\!:=\inf_{\tau\leq t, a\in\{0, 1\}}\pi_\tau(a|\tilde\xb_\tau, \cH_{\tau-1})$. Then under Assumptions \ref{ass:boundedness} and \ref{ass:min-signal}, there exist constants $C$ and $C_1'$ such that as long as $q_t\geq {C_1'}\max\left\{\frac{d(d+\log t)}{\lambda_0t}, \frac{\xi(d+\log t)}{\lambda_0^2t}\right\}$ and $\max_{a\in\{0, 1\}}\|\bDelta_t(a)\|_2\leq \frac{\lambda_{0}}{4d}$, with probability at least $1\!-\!\frac{8}{t^2}$,
\begin{equation}\label{eq:thm:theta-estimation-estimated-Sigma}
    \|\tilde\btheta_a^{(t)}-\btheta_a^*\|_2\leq \frac{C(R+R_{\btheta})d}{\lambda_{0}}\max\bigg\{\frac{d\!+\!\log t}{q_tt}, \frac{\sqrt{\xi}+\sqrt{\nu}}{\sqrt{d}}\cdot \sqrt{\frac{d\!+\!\log t}{q_tt}}\bigg\} + \frac{2R_\theta d}{\lambda_{0}}\|\bDelta_t(a)\|_2, \quad \forall a\in\{0, 1\}.
\end{equation}
\end{theorem}
The proof of Theorem \ref{thm:theta-estimation-estimated-Sigma} is in Appendix \ref{pf::theta-estimation-estimated-Sigma}.

By combining Theorem \ref{thm:theta-estimation-estimated-Sigma} and \ref{thm-regret} (with (\ref{eq::proposed-estimator-estimated-Sigma}) instead of (\ref{eq:proposed-estimator})), we obtain the following regret bounds for Algorithm \ref{alg1} with (\ref{eq::proposed-estimator-estimated-Sigma}) as the plug-in estimator.

\begin{corollary}\label{cor:alg1-with-theta-estimation-estimated-Sigma}
Suppose Assumption \ref{ass:boundedness} to \ref{ass:small-error} hold. Then there exist universal constants $C, C''$ such that:
\begin{itemize}
    \item[(i)] In the standard setting, if $\max_{t\in[dT^{2/3}, T]}\max_{a\in\{0, 1\}}\|\bDelta_t(a)\|_2\leq \frac{\lambda_0}{4d}$, $T\geq C''\max\{(1+{1}/{\lambda_0^{\frac94}})(d+\log T)^3, (\xi/\lambda_0)^{\frac94}(d+\log T)^{\frac43}\}$, by choosing $T_0 = \lceil2dT^{\frac23}\rceil$, $p_0^{(t)} = \min\{\frac12, t^{-\frac13}\}$, and (\ref{eq::proposed-estimator-estimated-Sigma}) instead of (\ref{eq:proposed-estimator}) in Algorithm \ref{alg1}, with probability at least $1-\frac{16}{\sqrt{T}}$,
    $$
\text{Regret}(T;\pi^*)\leq CdT^{\frac23}\left\{\frac{R_\theta}{1-\rho} + \frac{(\!\sqrt{\nu}\! +\!\! \sqrt{\xi}\! +\!\! 1)(R+R_\theta)}{(1-\rho)\lambda_0} \sqrt{1+\frac{\log T}{d}}\right\} + \frac{4R_\theta d}{(1-\rho)\lambda_0}\sum\limits_{t=T_0-1}^T\max\limits_{a\in\{0, 1\}}\|\bDelta_{t}(a)\|_2.
$$
    
\item[(ii)] In the clipped policy setting, as long as $\max_{t\in[d\sqrt{T}, T]}\max_{a\in\{0, 1\}}\|\bDelta_t(a)\|_2\leq \frac{\lambda_0}{4d}$, $T\geq C''\max\{(d+\log T)^2/(\lambda_0p_0)^2, \xi^2/\lambda_0^4(1+\log T/d)^2\}$, by choosing $T_0 = \lceil2d\sqrt{T}\rceil$, $p_0^{(t)}\equiv p_0$, and (\ref{eq::proposed-estimator-estimated-Sigma}) instead of (\ref{eq:proposed-estimator}) in Algorithm \ref{alg1}, with probability at least $1-\frac{16}{\sqrt{T}}$, 
$$
\text{Regret}(T;\bar\pi^*)\leq CdT^{\frac12}\bigg\{\!\!R_\theta + \frac{(\!\sqrt{\nu}\! +\!\! \sqrt{\xi}\! +\!\! 1)(1\!-\!2p_0)(R\!+\!R_\theta)}{\sqrt{p_0}(1\!-\!\rho)\lambda_0}\sqrt{1\!+\!\frac{\log T}{d}}\bigg\}+\frac{4(1\!-\!2p_0)R_\theta d}{(1-\rho)\lambda_0}\sum\limits_{t = T_0-1}^T\max\limits_{a\in\{0, 1\}}\|\bDelta_{t}(a)\|_2.
$$
\end{itemize}
\end{corollary}

The proof is in Appendix \ref{pf::cor:alg1-with-theta-estimation-estimated-Sigma}.

\section{Additional proofs}

\subsection{Proof of Theorem \ref{thm:regret-multiple-actions}}\label{pf:thm:regret-multiple-actions}

The proof of Theorem \ref{thm:regret-multiple-actions} is very similar to Theorem \ref{thm-regret}, and we only need to note the difference in Lemma \ref{lem::standard-setting-regret-hat} and \ref{lem::clipped-setting-regret-hat} (for the standard setting and the clipped policy setting respectively), as stated below. Recall that 
\begin{align*}
\widehat{\text{Regret}}_t(\pi, \pi^*) : = \EE_{a\sim \pi_t^*}\langle \btheta^*_{a}, \tilde\xb_t\rangle - \EE_{a\sim \pi_t}\langle \btheta^*_{a}, \tilde\xb_t\rangle, \\
\widehat{\text{Regret}}_t(\pi, \bar\pi^*) : = \EE_{a\sim \bar\pi_t^*}\langle \btheta^*_{a}, \tilde\xb_t\rangle - \EE_{a\sim \pi_t}\langle \btheta^*_{a}, \tilde\xb_t\rangle.
\end{align*}

\noindent\textbf{Standard setting.} At any time $t>T_0$, we have
\begin{align}
\widehat{\text{Regret}}_t(\pi, \pi^*) &= \EE_{a\sim \pi_t^*}\langle \btheta^*_{a}, \tilde\xb_t\rangle - \EE_{a\sim \pi_t}\langle \btheta^*_{a}, \tilde\xb_t\rangle\nonumber\\
&=\langle\btheta^*_{a_t^*}, \tilde\xb_t\rangle-\big[(1-(K-1)p_0^{(t)})\langle\btheta^*_{\tilde a_t}, \tilde\xb_t\rangle+\sum_{a\neq \tilde a_t}p_0^{(t)}\langle\btheta^*_{a}, \tilde\xb_t\rangle\big]\nonumber\\
&= p_0^{(t)}\sum_{a\neq a_t^*}\langle \btheta^*_{a_t^*} - \btheta^*_{a}, \tilde\xb_t\rangle + 1_{\{a_t^*\neq \tilde a_t\}}(1-Kp_0^{(t)})\langle \btheta^*_{a_t^*} - \btheta^*_{\tilde a_t}, \tilde\xb_t\rangle\nonumber\\
&\leq 2(K-1)p_0^{(t)}R_\theta + 1_{\{a_t^*\neq \tilde a_t\}}\langle \btheta^*_{a_t^*} - \btheta^*_{\tilde a_t}, \tilde\xb_t\rangle.\label{eq:instant-regret-standard-benchmark-multiple-actions}
\end{align}
Here note that Lemma \ref{lem::at-star=at-dagger} still holds under Assumption \ref{ass:small-error-K-actions}, so $a_t^*=\argmax_a\langle \btheta^*_{a}, \xb_t\rangle=\argmax_a\langle \btheta^*_{a}, \tilde\xb_t\rangle$, $\tilde a_t = \argmax_{a}\langle\hat\btheta_{a}^{(t-1)}, \tilde\xb_t\rangle$.

Note that $a_t^*\neq \tilde a_t$ implies that 
\begin{align*}
\begin{cases}
\langle\btheta_{a_t^*}^*, \tilde\xb_t\rangle\geq \langle\btheta_{\tilde a_t}^*, \tilde\xb_t\rangle\\
\langle\hat\btheta_{a_t^*}^{(t-1)}, \tilde\xb_t\rangle\leq \langle\hat\btheta_{\tilde a_t}^{(t-1)}, \tilde\xb_t\rangle
\end{cases}
\end{align*}
which leads to 
\begin{align*}
\langle \btheta_{a_t^*}^*, \tilde\xb_t\rangle&\geq \langle\btheta_{\tilde a_t}^*, \tilde\xb_t\rangle\geq \langle\hat\btheta_{\tilde a_t}^{(t-1)}, \tilde\xb_t\rangle - \max_a\|\hat\btheta_a^{(t-1)}-\btheta_a^*\|_2\\
&\geq \langle\hat\btheta_{a_t^*}^{(t-1)}, \tilde\xb_t\rangle - \max_a\|\hat\btheta_a^{(t-1)}-\btheta_a^*\|_2\geq \langle\btheta_{a_t^*}^*, \tilde\xb_t\rangle - 2\max_a\|\hat\btheta_a^{(t-1)}-\btheta_a^*\|_2,
\end{align*}
and further implies $\left|\langle \btheta^*_{a_t^*} - \btheta^*_{\tilde a_t}, \tilde\xb_t\rangle\right|\leq 2\max_a\|\hat\btheta_a^{(t-1)}-\btheta_a^*\|_2$.
Plugging in the above to (\ref{eq:instant-regret-standard-benchmark-multiple-actions}) leads to 
\begin{align*}
\widehat{\text{Regret}}_t(\pi, \pi^*)&\leq 2(K-1)p_0^{(t)}R_\theta + 2\max_a\|\hat\btheta_a^{(t-1)}-\btheta_a^*\|_2.
\end{align*}

The rest of the proof can be done in the same way as the proof of Theorem \ref{thm-regret}.

\noindent\textbf{Clipped policy setting.} At any time $t>T_0$, we have 
\begin{align}
\widehat{\text{Regret}}_t(\pi, \bar\pi^*) &= \EE_{a\sim \pi_t^*}\langle \btheta^*_{a}, \tilde\xb_t\rangle - \EE_{a\sim \pi_t}\langle \btheta^*_{a}, \tilde\xb_t\rangle\nonumber\\
&\leq (1-Kp_0)1_{\{a_t^*\neq \tilde a_t\}}\left|\langle \btheta^*_{a_t^*} - \btheta^*_{\tilde a_t}, \tilde\xb_t\rangle\right|.\label{eq:instant-regret-multiple-actions}
\end{align}
Here recall that $a_t^*=\argmax_a\langle \btheta^*_{a}, \tilde\xb_t\rangle$, $\tilde a_t := \argmax_{a}\langle\hat\btheta_{a}^{(t-1)}, \tilde\xb_t\rangle$.

Note that $a_t^*\neq \tilde a_t$ implies that 
\begin{align*}
\begin{cases}
\langle\btheta_{a_t^*}^*, \tilde\xb_t\rangle\geq \langle\btheta_{\tilde a_t}^*, \tilde\xb_t\rangle\\
\langle\hat\btheta_{a_t^*}^{(t-1)}, \tilde\xb_t\rangle\leq \langle\hat\btheta_{\tilde a_t}^{(t-1)}, \tilde\xb_t\rangle
\end{cases}
\end{align*}
which leads to 
\begin{align*}
\langle \btheta_{a_t^*}^*, \tilde\xb_t\rangle&\geq \langle\btheta_{\tilde a_t}^*, \tilde\xb_t\rangle\geq \langle\hat\btheta_{\tilde a_t}^{(t-1)}, \tilde\xb_t\rangle - \max_a\|\hat\btheta_a^{(t-1)}-\btheta_a^*\|_2\\
&\geq \langle\hat\btheta_{a_t^*}^{(t-1)}, \tilde\xb_t\rangle - \max_a\|\hat\btheta_a^{(t-1)}-\btheta_a^*\|_2\geq \langle\btheta_{a_t^*}^*, \tilde\xb_t\rangle - 2\max_a\|\hat\btheta_a^{(t-1)}-\btheta_a^*\|_2,
\end{align*}
and further implies $\left|\langle \btheta^*_{a_t^*} - \btheta^*_{\tilde a_t}, \tilde\xb_t\rangle\right|\leq 2\max_a\|\hat\btheta_a^{(t-1)}-\btheta_a^*\|_2$.

Plugging in the above to (\ref{eq:instant-regret-multiple-actions}) leads to 
$$
\widehat{\text{Regret}}_t(\pi, \bar\pi^*)\leq 2(1-Kp_0)\max_a\|\hat\btheta_a^{(t-1)}-\btheta_a^*\|_2
$$
for $t>T_0$. The rest of the proof can be done in the same way as the proof of Theorem \ref{thm-regret}.

\subsection{Proof of Lemma \ref{lem:hat-Sigma-x-r}}\label{pf:lem:hat-Sigma-x-r}
We first analyze $\hat\bSigma_{\tilde\xb, a}^{(t)}$. Notice that $\hat\bSigma_{\tilde\xb, a}^{(t)}=\frac{1}{t}\sum_{\tau\in[t]}\bV_{\tau, a}$, where $\bV_{\tau, a}=\frac{\pi_{\tau}^{nd}(A_\tau)}{\pi_\tau(A_\tau|\tilde\xb_\tau, \cH_{\tau-1})}1_{\{A_\tau=a\}}\tilde\xb_\tau\tilde\xb_{\tau}^\top$.  For any fixed $\bu\in\mathbb S^{d-1}:= \{\bu'\in \mathbb R^d: \|\bu'\|_2=1\}$, $(v_{\bu, \tau, a}:=\bu^\tau[\bV_{\tau, a}-\EE[\bV_{\tau, a}|\cH_{\tau-1}]]\bu)_\tau$ is a martingale difference sequence. Moreover, we can verify that $|v_{\bu, \tau, a}|\leq \frac{2}{q_t}$ and
\begin{align*}
    \Var(v_{\bu, \tau, a}|\cH_{\tau-1})&
    \leq \EE[(\bu^\top \bV_{\tau, a}\bu)^2|\cH_{\tau-1}]=\EE\left[\left(\frac{\pi_{\tau}^{nd}(A_\tau)}{\pi_\tau(A_\tau|\tilde\xb_\tau, \cH_{\tau-1})}\right)^21_{\{A_\tau=a\}}(\bu^\top\tilde\xb_\tau\tilde\xb_\tau^\top\bu)^2\bigg|\cH_{\tau-1}\right]\\
    &=\EE_{\bepsilon_\tau, A_\tau\sim \pi_{\tau}^{nd}(\cdot)}\left[\left(\frac{\pi_{\tau}^{nd}(A_\tau)}{\pi_\tau(A_\tau|\tilde\xb_\tau, \cH_{\tau-1})}\right)1_{\{A_\tau=a\}}(\bu^\top\tilde\xb_\tau\tilde\xb_\tau^\top\bu)^2\bigg|\cH_{\tau-1}\right]\leq \frac{1}{q_t}\mathbb E(\bu^\top\tilde\xb_\tau)^4\leq \frac{\xi}{d^2q_t}.
\end{align*}

According to Freedman's Inequality \citep{freedman1975tail}, for any $\gamma_1, \gamma_2>0$, 
$$
\PP\left(\sum_{\tau\in[t]}v_{\bu, \tau, a}\geq \gamma_1, \sum_{\tau\in[t]}\Var(v_{\bu, \tau, a}|\cH_{\tau-1})\leq \gamma_2\right)\leq e^{-\frac{\gamma_1^2}{2(\frac{2}{q_t}\gamma_1+\gamma_2)}}.
$$
Set $\gamma_2 = \frac{\xi t}{d^2q_t}$, and we obtain 
$\PP\left(\sum_{\tau\in[t]}v_{\bu, \tau, a}\geq \gamma_1\right)\leq e^{-\frac{d^2q_t\gamma_1^2}{2(2d^2\gamma_1+\xi t)}}
$. Applying the same analysis to $(-v_{\bu, \tau, a})_\tau$ and combining the results gives $\PP\left(|\sum_{\tau\in[t]}v_{\bu, \tau, a}|\geq \gamma_1\right)\leq 2e^{-\frac{d^2q_t\gamma_1^2}{2(2d^2\gamma_1+\xi t)}}
$.

Denote $\bM_t= \frac{1}{t}\sum_{\tau\in[t]}(\bV_{\tau, a}-\EE[\bV_{\tau, a}|\cH_{\tau-1}])$, then the above means that $\forall \bu\in\mathbb S^{d-1}$, 
\begin{equation}\label{eq:v-concentration-singledirection} 
  \PP\left(|\bu^\top\bM_t\bu|\geq \frac{\gamma_1}{t}\right)\leq 2e^{-\frac{d^2q_t\gamma_1^2}{2(2d^2\gamma_1+\xi t)}}.
\end{equation}%
Let $\cN$ be a $\frac{1}{4}$-net of $\mathbb S^{d-1}$, $|\cN|\leq 9^d$. $\forall \bu\in\mathbb S^{d-1}$, find $\bu'\in\cN$ s.t. $\|\bu-\bu'\|_2\leq \frac{1}{4}$, and we have 
$$
|\bu^\top \bM_t\bu-\bu'^\top\bM_t\bu'|\leq |\bu^\top \bM_t(\bu-\bu')| + |\bu'^\top\bM_t(\bu-\bu')|\leq \frac12 \|\bM_t\|_2.
$$
This implies that 
$$
\|\bM_t\|_2 = \sup_{\bu\in\mathbb S^{d-1}}|\bu^\top\bM_t\bu|\leq \sup_{\bu'\in\mathcal \cN}|\bu'^\top\bM_t\bu'|+\frac12 \|\bM_t\|_2,
$$
and thus $\sup_{\bu\in\cN}|\bu^\top\bM_t\bu|\geq\frac12\|\bM_t\|_2$.
Combining the above and (\ref{eq:v-concentration-singledirection}), we obtain that for any $\gamma_1>0$,
\begin{align*}
\PP\left(\|\bM_t\|_2\geq \frac{2\gamma_1}{t}\right)\leq \PP\left(\sup_{\bu\in\cN}|\bu^\top\bM_t\bu|\geq \frac{\gamma_1}{t}\right)\leq 9^d\cdot \PP\left(|\bu^\top\bM_t\bu|\geq \frac{\gamma_1}{t}\right)=2\cdot 9^d\cdot e^{-\frac{d^2q_t\gamma_1^2}{2(2d^2\gamma_1+\xi t)}}.
\end{align*}
By choosing $\gamma_1 = 24\max\{\frac{d+\log t}{q_t}, \frac{\sqrt{\xi}}{d}\cdot \sqrt{\frac{t(d+\log t)}{q_t}}\}$, and noticing that \\$\bM_t = \hat\bSigma_{\tilde\xb, a}^{(t)}-\frac{1}{t}\sum_{\tau\in[t]}\EE[\bV_{\tau, a}|\cH_{\tau-1}]$, we have
\begin{equation}\label{eq:hat-Sigma-x-1}
\PP\left(\bigg\|\hat\bSigma_{\tilde\xb, a}^{(t)}-\frac{1}{t}\sum_{\tau\in[t]}\EE[\bV_{\tau, a}|\cH_{\tau-1}]\bigg\|_2\geq 48\max\left\{\frac{d+\log t}{q_t t}, \frac{\sqrt{\xi}}{d}\cdot \sqrt{\frac{d+\log t}{q_t t}}\right\}\right)\leq \frac{2}{t^2}.
\end{equation}
At the same time, we have 
\begin{align}
\EE[\bV_{\tau, a}|\cH_{\tau-1}]
&= \EE_{\bepsilon_\tau}\left[\EE_{A_\tau\sim\pi_\tau(\cdot|\tilde\xb_{\tau}, \cH_{\tau-1})}\bigg[\frac{\pi_{\tau}^{nd}(A_\tau)}{\pi_\tau(A_\tau|\tilde\xb_\tau, \cH_{\tau-1})}1_{\{A_\tau=a\}}\tilde\xb_\tau\tilde\xb_{\tau}^\top\big|\bepsilon_\tau, \cH_{\tau-1}\bigg]\bigg|\cH_{\tau-1}\right]\nonumber\\
& =\EE_{\bepsilon_\tau}\left[\EE_{A_\tau\sim\pi_{\tau}^{nd}(\cdot)}\big[1_{\{A_\tau=a\}}\tilde\xb_\tau\tilde\xb_{\tau}^\top\big|\bepsilon_\tau, \cH_{\tau-1}\big]\bigg|\cH_{\tau-1}\right]\nonumber\\
& =\EE_{\bepsilon_\tau}\left[\pi_{\tau}^{nd}(a)\cdot\tilde\xb_\tau\tilde\xb_{\tau}^\top\bigg|\cH_{\tau-1}\right]\nonumber\\
&=\pi_{\tau}^{nd}(a)(\xb_\tau\xb_{\tau}^\top+\bSigma_{e, \tau}).\label{eq:condition-e-V}
\end{align}
Here we've used the facts that (i) $\{\pi_{\tau}^{nd}\}_{\tau}$ is data-independent; (ii) $\EE[\bepsilon_\tau|\cH_{\tau-1}] = 0$, $\Var[\bepsilon_\tau|\cH_{\tau-1}] = \bSigma_{e, \tau}$.
Plug (\ref{eq:condition-e-V}) into (\ref{eq:hat-Sigma-x-1}), and we get
\begin{equation}\label{eq:hat-Sigma-x-2}
\PP\left(\bigg\|\hat\bSigma_{\tilde\xb, a}^{(t)}-\frac{1}{t}\sum_{\tau\in[t]}\pi_{\tau}^{nd}(a)(\xb_\tau\xb_{\tau}^\top+\bSigma_{e, \tau})\bigg\|_2\geq 48\max\left\{\frac{d+\log t}{q_t t}, \frac{\sqrt{\xi}}{d}\cdot \sqrt{\frac{d+\log t}{q_t t}}\right\}\right)\leq \frac{2}{t^2}.
\end{equation}

The analysis for $\hat\bSigma_{\tilde\xb, r, a}^{(t)}$ is similar. Write $\hat\bSigma_{\tilde\xb, r, a}^{(t)}=\frac{1}{t}\sum_{\tau\in[t]}\bZ_{\tau, a}$, $\bZ_{\tau, a}:=\frac{\pi_{\tau}^{nd}(A_\tau)}{\pi_\tau(A_\tau|\tilde\xb_\tau, \cH_{\tau-1})}1_{\{A_\tau = a\}}\tilde\xb_\tau r_\tau$. Then for any $\bu\in \mathcal S^{d-1}$, it's easy to verify that $|(\bZ_{\tau, a}-\EE[\bZ_{\tau, a}|\cH_{\tau-1}])^\top\bu|\leq \frac{2R}{q_t}$, and $\Var((\bZ_{\tau, a}-\EE[\bZ_{\tau, a}|\cH_{\tau-1}])^\top\bu|\cH_{\tau - 1})\leq \EE[(\bZ_{\tau, a}^\top\bu)^2|\cH_{\tau-1}]\leq\frac{\nu R^2}{q_td}$. Applying Freedman's Inequality leads to 
\begin{equation}\label{eq:z-concentration-singledirection}
\PP\left(|\bz_t^\top \bu|\geq \frac{\gamma_1}{t}\right)\leq 2e^{-\frac{dq_t\gamma_1^2}{4Rd\gamma_1+2R^2\nu t}},
\end{equation}
where $\bz_t:=\frac{1}{t}\sum_{\tau\in[t]}(\bZ_{\tau, a}-\EE[\bZ_{\tau, a}|\cH_{\tau-1}])$. 

Recall that $\cN$ is a $\frac{1}{4}$-net of $\mathbb S^{d-1}$, $|\cN|\leq 9^d$. $\forall \bu\in\mathbb S^{d-1}$, find $\bu'\in\mathbb S^{d-1}$ s.t. $\|\bu-\bu'\|\leq 1/4$, then $|\bz_t^\top\bu-\bz_t^\top \bu'|\leq \frac{1}{4}\|\bz_t\|_2$, and thus 
$$
\|\bz_t\|_2=\sup_{\bu\in\mathbb S^{d-1}}|\bz_t^\top\bu|\leq \sup_{\bu'\in\cN}|\bz_t^\top\bu'|+\frac14 \|\bz_t\|_2
$$
which implies that $\sup_{\bu\in\cN}|\bz_t^\top \bu|\geq \frac34 \|\bz_t\|_2$. Taking this and (\ref{eq:z-concentration-singledirection}) into account, we derive that
$$
\PP\left(\|\bz_t\|_2\geq \frac{4}{3}\frac{\gamma_1}{t}\right)\leq \PP\left(\sup_{\bu\in\cN}|\bz_t^\top\bu|\geq \frac{\gamma_1}{t}\right)\leq 9^d\cdot 2e^{-\frac{dq_t\gamma_1^2}{4Rd\gamma_1+2R^2\nu t}}
$$
By choosing $\gamma_1=24R\max\left\{\frac{d+\log t}{q_t}, \sqrt{\nu}\cdot \sqrt{\frac{(d+\log t)t}{dq_t}}\right\}$ and noticing that $\bz_t = \hat\bSigma_{\tilde\xb, r, a}^{(t)}-\frac{1}{t}\sum_{\tau\in[t]}\EE[\bZ_{\tau, a}|\cH_{\tau-1}]$, we obtain 
\begin{equation}\label{eq:hat-Sigma-x-r-1}
\PP\left(\bigg\|\hat\bSigma_{\tilde\xb, r, a}^{(t)}-\frac{1}{t}\sum_{\tau\in[t]}\EE[\bZ_{\tau, a}|\cH_{\tau-1}]\bigg\|_2\geq 32R\max\left\{\frac{d+\log t}{q_tt}, \sqrt{\frac{\nu}{d}}\cdot \sqrt{\frac{d+\log t}{q_tt}}\right\}\right)\leq \frac{2}{t^2}.
\end{equation}
Finally, because 
\begin{align*}
\EE[\bZ_{\tau, a}|\cH_{\tau-1}]
&=\EE_{\bepsilon_\tau}\left[\EE_{A_\tau\sim\pi_\tau(\cdot|\tilde\xb_\tau, \cH_{\tau-1}), \eta_\tau}\left[\frac{\pi_{\tau}^{nd}(A_\tau)}{\pi_{\tau}(A_\tau|\tilde\xb_\tau, \cH_{\tau-1})}1_{\{A_\tau=a\}}\tilde\xb_\tau r_\tau\big|\cH_{\tau-1}, \bepsilon_\tau\right]\bigg|\cH_{\tau-1}\right]\\
&=\EE_{\bepsilon_\tau}\left[\EE_{A_\tau\sim\pi_{\tau}^{nd}(\cdot), \eta_\tau}\left[1_{\{A_\tau=a\}}\tilde\xb_\tau r_\tau|\cH_{\tau-1}, \bepsilon_\tau\right]\bigg|\cH_{\tau-1}\right]\\
&=\EE_{\bepsilon_\tau}\left[\EE_{A_\tau\sim\pi_{\tau}^{nd}(\cdot), \eta_\tau}\left[1_{\{A_\tau=a\}}\tilde\xb_\tau (\xb_{\tau}^\top\btheta^*_{a}+\eta_\tau)|\cH_{\tau-1}, \bepsilon_\tau\right]\bigg|\cH_{\tau-1}\right]\\
&=\EE_{\bepsilon_\tau}\left[\pi_{\tau}^{nd}(a)\EE_{\eta_\tau}\left[\tilde\xb_\tau (\xb_{\tau}^\top\btheta^*_{a}+\eta_\tau)|\cH_{\tau-1}, \bepsilon_\tau\right]\bigg|\cH_{\tau-1}\right]\\
&=\pi_{\tau}^{nd}(a)\EE_{\bepsilon_\tau}\left[\tilde\xb_\tau \xb_{\tau}^\top\btheta^*_{a}\bigg|\cH_{\tau-1}\right]=\pi_{\tau}^{nd}(a)(\xb_\tau \xb_{\tau}^\top)\btheta^*_{a},
\end{align*}
Plug in (\ref{eq:hat-Sigma-x-r-1}), and we obtain 
\begin{equation}\label{eq:hat-Sigma-x-r-2}
\PP\left(\bigg\|\hat\bSigma_{\tilde\xb, r, a}^{(t)}-\bigg[\frac{1}{t}\sum_{\tau\in[t]}\pi_{\tau}^{nd}(a)\xb_\tau \xb_{\tau}^\top\bigg]\btheta^*_{a}\bigg\|_2\geq 32R\max\left\{\frac{d+\log t}{q_tt}, \sqrt{\frac{\nu}{d}}\cdot \sqrt{\frac{d+\log t}{q_tt}}\right\}\right)\leq \frac{2}{t^2}.
\end{equation}
Combining (\ref{eq:hat-Sigma-x-r-2}) and (\ref{eq:hat-Sigma-x-2}), we conclude the proof.

\subsection{Proof of Lemma \ref{lem::at-star=at-dagger}}\label{pf::lem::at-star=at-dagger}

We only need to prove
\begin{equation}\label{eq::at-star=at-dagger}
\sign(\langle \btheta_1^* - \btheta_0^*, \xb_t\rangle) = \sign(\langle \btheta_1^* - \btheta_0^*, \tilde\xb_t\rangle).
\end{equation}
If $\langle\btheta_1^* - \btheta_0^*, \xb_t \rangle = 0$, (\ref{eq::at-star=at-dagger}) is a direct consequence of Assumption \ref{ass:small-error}. If $\langle\btheta_1^* - \btheta_0^*, \xb_t \rangle \neq 0$, without loss of generality, suppose $\langle\btheta_1^* - \btheta_0^*, \xb_t \rangle > 0$. 
Then according to Assumption \ref{ass:small-error},
$$
\langle\btheta_1^* - \btheta_0^*, \tilde\xb_t \rangle = \langle\btheta_1^* - \btheta_0^*, \xb_t \rangle + \langle\btheta_1^* - \btheta_0^*, \bepsilon_t \rangle\geq(1-\rho)\langle\btheta_1^* - \btheta_0^*, \xb_t \rangle>0.
$$
Thus (\ref{eq::at-star=at-dagger}) is true.

\subsection{Proof of Lemma \ref{lem::standard-setting-regret-hat}}\label{pf::lem::standard-setting-regret-hat}

{
We have 
\begin{align}
\widehat{\text{Regret}}_t(\pi, \pi^*) &= \EE_{a\sim \pi_t^*}\langle \btheta^*_{a}, \tilde\xb_t\rangle - \EE_{a\sim \pi_t}\langle \btheta^*_{a}, \tilde\xb_t\rangle\nonumber\\
&=\langle\btheta^*_{a_t^*}, \tilde\xb_t\rangle-[(1-p_0^{(t)})\langle\btheta^*_{\tilde a_t}, \tilde\xb_t\rangle+p_0^{(t)}\langle\btheta^*_{1-\tilde a_t}, \tilde\xb_t\rangle]\nonumber\\
&= 1_{\{a_t^*= \tilde a_t\}}p_0^{(t)}\langle \btheta^*_{a_t^*} - \btheta^*_{1-a_t^*}, \tilde\xb_t\rangle + 1_{\{a_t^*\neq \tilde a_t\}}(1-p_0^{(t)})\langle \btheta^*_{a_t^*} - \btheta^*_{1-a_t^*}, \tilde\xb_t\rangle\nonumber\\
&\leq 2p_0^{(t)}R_\theta + 1_{\{a_t^*\neq \tilde a_t\}}\langle \btheta^*_{a_t^*} - \btheta^*_{1-a_t^*}, \tilde\xb_t\rangle.\label{eq:instant-regret-standard-benchmark}
\end{align}

Here recall that $a_t^*=\argmax_a\langle \btheta^*_{a}, \xb_t\rangle = \argmax_a\langle \btheta^*_{a}, \tilde\xb_t\rangle$, $\tilde a_t := \argmax_{a\in\{0, 1\}}\langle\hat\btheta_{a}^{(t-1)}, \tilde\xb_t\rangle$.

Note that $a_t^*\neq \tilde a_t$ implies that 
\begin{align*}
\begin{cases}
\langle\btheta_{a_t^*}^*, \tilde\xb_t\rangle\geq \langle\btheta_{1-a_t^*}^*, \tilde\xb_t\rangle\\
\langle\hat\btheta_{a_t^*}^{(t-1)}, \tilde\xb_t\rangle\leq \langle\hat\btheta_{1-a_t^*}^{(t-1)}, \tilde\xb_t\rangle
\end{cases}
\end{align*}
which leads to 
\begin{align*}
\langle \btheta_{a_t^*}^*, \tilde\xb_t\rangle&\geq \langle\btheta_{1-a_t^*}^*, \tilde\xb_t\rangle\geq \langle\hat\btheta_{1-a_t^*}^{(t-1)}, \tilde\xb_t\rangle - \max_a\|\hat\btheta_a^{(t-1)}-\btheta_a^*\|_2\\
&\geq \langle\hat\btheta_{a_t^*}^{(t-1)}, \tilde\xb_t\rangle - \max_a\|\hat\btheta_a^{(t-1)}-\btheta_a^*\|_2\geq \langle\btheta_{a_t^*}^*, \tilde\xb_t\rangle - 2\max_a\|\hat\btheta_a^{(t-1)}-\btheta_a^*\|_2,
\end{align*}
and further implies $\left|\langle \btheta^*_{a_t^*} - \btheta^*_{1-a_t^*}, \tilde\xb_t\rangle\right|\leq 2\max_a\|\hat\btheta_a^{(t-1)}-\btheta_a^*\|_2$.

Plugging in the above to (\ref{eq:instant-regret-standard-benchmark}) leads to 
$$
\widehat{\text{Regret}}_t(\pi, \pi^*)\leq 2p_0^{(t)}R_\theta + 2\max_a\|\hat\btheta_a^{(t-1)}-\btheta_a^*\|_2.
$$
}

\subsection{Proof of Lemma \ref{lem::clipped-setting-regret-hat}}\label{pf::lem::clipped-setting-regret-hat}

{
At any time $t>T_0$, we have 
\begin{align}
\widehat{\text{Regret}}_t(\pi, \bar\pi^*) &= \EE_{a\sim \bar\pi_t^*}\langle \btheta^*_{a}, \tilde\xb_t\rangle - \EE_{a\sim \pi_t}\langle \btheta^*_{a}, \tilde\xb_t\rangle\nonumber\\
&=[(1-p_0)\langle\btheta^*_{a_t^*}, \tilde\xb_t\rangle+p_0\langle\btheta^*_{1-a_t^*}, \tilde\xb_t\rangle]-[(1-p_0)\langle\btheta^*_{\tilde a_t}, \tilde\xb_t\rangle+p_0\langle\btheta^*_{1-\tilde a_t}, \tilde\xb_t\rangle]\nonumber\\
&\leq (1-2p_0)1_{\{a_t^*\neq \tilde a_t\}}\left|\langle \btheta^*_{a_t^*} - \btheta^*_{1-a_t^*}, \tilde\xb_t\rangle\right|.\label{eq:instant-regret}
\end{align}
Here recall that $a_t^*=\argmax_a\langle \btheta^*_{a}, \xb_t\rangle = \argmax_a\langle \btheta^*_{a}, \tilde\xb_t\rangle$, $\tilde a_t := \argmax_{a\in\{0, 1\}}\langle\hat\btheta_{a}^{(t-1)}, \tilde\xb_t\rangle$.

Note that $a_t^*\neq \tilde a_t$ implies that 
\begin{align*}
\begin{cases}
\langle\btheta_{a_t^*}^*, \tilde\xb_t\rangle\geq \langle\btheta_{1-a_t^*}^*, \tilde\xb_t\rangle\\
\langle\hat\btheta_{a_t^*}^{(t-1)}, \tilde\xb_t\rangle\leq \langle\hat\btheta_{1-a_t^*}^{(t-1)}, \tilde\xb_t\rangle
\end{cases}
\end{align*}
which leads to 
\begin{align*}
\langle \btheta_{a_t^*}^*, \tilde\xb_t\rangle&\geq \langle\btheta_{1-a_t^*}^*, \tilde\xb_t\rangle\geq \langle\hat\btheta_{1-a_t^*}^{(t-1)}, \tilde\xb_t\rangle - \max_a\|\hat\btheta_a^{(t-1)}-\btheta_a^*\|_2\\
&\geq \langle\hat\btheta_{a_t^*}^{(t-1)}, \tilde\xb_t\rangle - \max_a\|\hat\btheta_a^{(t-1)}-\btheta_a^*\|_2\geq \langle\btheta_{a_t^*}^*, \tilde\xb_t\rangle - 2\max_a\|\hat\btheta_a^{(t-1)}-\btheta_a^*\|_2,
\end{align*}
and further implies $\left|\langle \btheta^*_{a_t^*} - \btheta^*_{1-a_t^*}, \tilde\xb_t\rangle\right|\leq 2\max_a\|\hat\btheta_a^{(t-1)}-\btheta_a^*\|_2$.

Plugging in the above to (\ref{eq:instant-regret}) leads to 
$$
\widehat{\text{Regret}}_t(\pi, \bar\pi^*)\leq 2(1-2p_0)\max_a\|\hat\btheta_a^{(t-1)}-\btheta_a^*\|_2
$$
}
\subsection{Proof of Theorem \ref{thm:theta-estimation-estimated-Sigma}}\label{pf::theta-estimation-estimated-Sigma}

Fix $t\in[T]$ such that the conditions of Theorem 2.3 hold. Fix $a\in\{0, 1\}$. As in Appendix \ref{pf:thm:theta-estimation}, define $\bDelta_1 :=\hat\bSigma_{\tilde\xb, a}^{(t)}-\frac{1}{t}\sum_{\tau\in[t]}\pi_{\tau}^{nd}(a)(\xb_\tau\xb_{\tau}^\top+\bSigma_{e, \tau})$, $\bDelta_2 := \hat\bSigma_{\tilde\xb, r, a}^{(t)}-\bigg(\frac{1}{t}\sum_{\tau\in[t]}\pi_{\tau}^{nd}(a)\xb_\tau \xb_{\tau}^\top\bigg)\btheta^*_{a}$. We also let $\bDelta_3 := -\bDelta_t(a) = -\frac{1}{t}\sum_{\tau\in[t]}\pi_{\tau}^{nd}(a)(\hat\bSigma_{e, \tau} - \bSigma_{e, \tau})$. Recall Lemma \ref{lem:hat-Sigma-x-r}: with probability at least $1-4/t^2$, 
$$
\|\bDelta_1\|_2\leq C\max\left\{\frac{d+\log t}{q_tt}, \frac{\sqrt{\xi}}{d}\sqrt{\frac{d+\log t}{q_tt}}\right\},\quad \|\bDelta_2\|_2\leq CR\max\left\{\frac{d+\log t}{q_tt}, \sqrt{\frac{\nu}{d}}\sqrt{\frac{d+\log t}{q_tt}}\right\}.
$$
Meanwhile, 
\begin{align*}
\tilde\btheta_a^{(t)}&= \bigg(\hat\bSigma_{\tilde\xb, a}^{(t)} - 
\frac{1}{t}\sum_{\tau\in[t]}\pi^{nd}_\tau(a)\hat\bSigma_{e, \tau}\bigg)^{-1}\cdot
\hat\bSigma_{\tilde\xb, r, a}^{(t)}\\
& = \bigg(\frac{1}{t}\sum_{\tau\in[t]}\pi^{nd}_\tau(a)\xb_\tau\xb_{\tau}^\top+\bDelta_1+\bDelta_3\bigg)^{-1}\cdot \left[\bigg(\frac{1}{t}\sum_{\tau\in[t]}\pi_{\tau}^{nd}(a)\xb_\tau \xb_{\tau}^\top\bigg)\btheta^*_{a}+\bDelta_2\right]\\
&= \btheta_a^* - \bJ_1' + \bJ_2',
\end{align*}
where 
\begin{align*}
\bJ_1':=\bigg[\frac{1}{t}\sum_{\tau\in[t]}\pi_{\tau}^{nd}(a)\xb_\tau\xb_{\tau}^\top+\bDelta_1+\bDelta_3\bigg]^{-1}(\bDelta_1+\bDelta_3)\btheta_a^*,\thickspace
\bJ_2':=\bigg[\frac{1}{t}\sum_{\tau\in[t]}\pi_{\tau}^{nd}(a)\xb_\tau\xb_{\tau}^\top+\bDelta_1+\bDelta_3\bigg]^{-1}\bDelta_2.
\end{align*}
Under the events where both (\ref{eq:hat-Sigma-x}) and (\ref{eq:hat-Sigma-x-r}) hold, whenever 
\begin{equation}\label{eq:condition-eigv-control-estimated-Sigma}
C\max\left\{\frac{d+\log t}{q_tt}, \frac{\sqrt{\xi}}{d}\sqrt{\frac{d+\log t}{q_tt}}\right\}\leq \frac{\lambda_{0}}{4d}
\end{equation}
and $\|\bDelta_3\|_2\leq \frac{\lambda_{0}}{4d}$, we have 
$$
\|\bJ_1'\|_2\leq \frac{2dR_{\btheta}}{\lambda_{0}}\left(C\max\left\{\frac{d+\log t}{q_tt}, \frac{\sqrt{\xi}}{d}\sqrt{\frac{d+\log t}{q_tt}}\right\} + \|\bDelta_3\|_2\right)
$$
and 
$$
\|\bJ_2'\|_2\leq \frac{2CdR}{\lambda_{0}}\max\left\{\frac{d+\log t}{q_tt}, \sqrt{\frac{\nu}{d}}\sqrt{\frac{d+\log t}{q_tt}}\right\}.
$$
(\ref{eq:condition-eigv-control-estimated-Sigma}) can be ensured by $t\geq {C_1'}\max\left\{\frac{d(d+\log t)}{\lambda_0q_t}, \frac{\xi(d+\log t)}{\lambda_0^2q_t}\right\}$, where $C_1' = \max\{4C, 16C^2\}$. Given these guarantees, we have with probability at least $1-4/t^2$, 
$$
\|\tilde\btheta_a^{(t)}-\btheta_a^*\|\leq \|\bJ_1'\|_2+\|\bJ_2'\|_2\leq \frac{2C(R+R_{\btheta})d}{\lambda_{0}}\max\left\{\frac{d+\log t}{q_tt}, \frac{\sqrt{\xi} + \sqrt{\nu}}{\sqrt{d}}\sqrt{\frac{d+\log t}{q_tt}}\right\} + \frac{2R_\theta d}{\lambda_0}\|\bDelta_t(a)\|_2.
$$
Thus we conclude the proof.

\subsection{{Proof of Corollary \ref{cor:alg1-with-theta-estimation-estimated-Sigma}}}\label{pf::cor:alg1-with-theta-estimation-estimated-Sigma}

\noindent\textbf{Standard setting.} Notice that $p_0^{(t)}$ is monotonically decreasing in $t$. Theorem \ref{thm:theta-estimation-estimated-Sigma} indicates that, as long as $\forall t> T_0$,
\begin{equation}\label{eq:standard-benchmark-condition1-estimated-sigma}
p_0^{(t)}\geq {C'}\max\left\{\frac{d(d+\log t)}{\lambda_0 t}, \frac{\xi(d+\log t)}{\lambda_0^2t}\right\},
\end{equation}
then with probability at least $1-\frac{8}{t^2}$, $\forall a\in\{0, 1\}$,
\begin{align}
\|\hat\btheta_a^{(t)}-\btheta_a^*\|_2
&\leq \frac{C(R+\!R_\theta)d}{\lambda_0}\max\left\{\!\frac{d+\log t}{t^{\frac23}}, \frac{\sqrt{\nu} + \sqrt{\xi}}{\sqrt{d}}\sqrt{\frac{d+\log t}{t^{\frac23}}}\right\} + \frac{2R_\theta d}{\lambda_0}\|\bDelta_t(a)\|_2\label{eq:standard-benchmark-estimation-error-estimated-sigma}
\end{align}

Plug it into Theorem \ref{thm-regret}, we have that with high probability, 
\begin{align*}
\text{Regret}(T;\pi^*)
&\leq 2R_\theta\cdot \lceil 2dT^{2/3}\rceil + \frac2{1-\rho} I_1',
\end{align*}
where
\begin{align*}
I_1& = \sum_{t=T_0+1}^T
\bigg(p_0^{(t)}R_\theta + \max_a\|\tilde \btheta_a^{(t-1)} - \btheta^*_a\|_2\bigg)\\
&\leq \sum_{t=T_0+1}^T t^{-\frac13}R_\theta + \sum_{t=T_0}^{T-1}\left[\frac{C(R+\!R_\theta)d}{\lambda_0}\max\left\{\!\frac{d+\log t}{t^{\frac23}}, \frac{\sqrt{\nu} + \sqrt{\xi}}{\sqrt{d}}\sqrt{\frac{d+\log t}{t^{\frac23}}}\right\}+\max_a\frac{2 R_\theta d}{\lambda_0}\|\bDelta_t(a)\|_2\right]\\
&\leq 2R_\theta T^{\frac23} +\frac{C'}{\lambda_0}(\sqrt{\nu} + \sqrt{\xi} + 1)(R+\!R_\theta)\sqrt{d(d+\log T)} T^{\frac23} + \frac{2 R_\theta d}{\lambda_0}\sum_{t\geq T_0}\|\bDelta_t(a)\|_2,
\end{align*}
for a universal constant $C'$, where the last inequality holds if in addition, $T\geq[d(d+\log T)]^{\frac32}$.

The proof is concluded by combining the above requirement for $T$ as well as \eqref{eq:standard-benchmark-condition1-estimated-sigma}.

\noindent\textbf{Clipped policy setting.} 

Similar to the standard setting, according to Theorem \ref{thm:theta-estimation-estimated-Sigma}, as long as $\forall t> T_0$,
\begin{equation}\label{eq:clipped-benchmark-condition1-estimated-sigma}
p_0\geq {C_1}\max\left\{\frac{d(d+\log t)}{\lambda_0 t}, \frac{\xi(d+\log t)}{\lambda_0^2t}\right\},
\end{equation}
then with probability at least $1-\frac{8}{t^2}$, $\forall a\in\{0, 1\}$,
\begin{align}
\|\tilde\btheta_a^{(t)}-\btheta_a^*\|_2
&\leq \frac{C(R+\!R_\theta)d}{\lambda_0}\max\left\{\!\frac{d+\log t}{p_0 t}, \frac{\sqrt{\nu} + \sqrt{\xi}}{\sqrt{d}}\sqrt{\frac{d+\log t}{p_0 t}}\right\} + \frac{2R_\theta d}{\lambda_0}\|\bDelta_t(a)\|_2\label{eq:clipped-benchmark-estimation-error-estimated-sigma}
\end{align}

Plug it into Theorem \ref{thm-regret}, we have that with high probability, 
\begin{align*}
\text{Regret}(T;\bar \pi^*)
&\leq 2R_\theta\cdot \lceil 2dT^{1/2}\rceil + \frac{2(1-2p_0)}{1-\rho} I_2',
\end{align*}
where
\begin{align*}
I_2'& = \sum_{t=T_0+1}^T \max_a\|\tilde \btheta_a^{(t-1)}-\btheta^*_a\|_2\\
&\leq \sum_{t=T_0}^{T-1}\left[\frac{C(R+\!R_\theta)d}{\lambda_0}\max\left\{\!\frac{d+\log t}{p_0 t}, \frac{\sqrt{\nu} + \sqrt{\xi}}{\sqrt{d}}\sqrt{\frac{d+\log t}{p_0 t}}\right\} + \max_a\frac{2R_\theta d}{\lambda_0}\|\bDelta_t(a)\|_2\right]\\
&\leq \frac{2C(R+R_\theta)(\sqrt{\nu}+\sqrt{\xi}+1)}{\lambda_0\sqrt{p_0}}\sqrt{d(d+\log T)}\sqrt{T} + \frac{2R_\theta d}{\lambda_0}\sum_{t=T_0}^{T-1}\max_a\|\bDelta_t(a)\|_2,
\end{align*}
where the last inequality holds if in addition, $T\geq d(d+\log T)\log^2 T$. 

The proof is concluded by plugging the above into the regret upper bound formula and combining the requirements for $T$.

\vfill





\end{document}